\documentclass{egpubl}
\usepackage{sca2024}

\SpecialIssuePaper         % uncomment for final version of Computer Graphics Forum, special issue
\CGFccby
\usepackage[T1]{fontenc}
\usepackage{dfadobe}  

\usepackage{cite}  % comment out for biblatex with backend=biber

\usepackage{float}
\usepackage{booktabs}
\usepackage{multirow}
\usepackage{makecell}
\usepackage{multicol}
\usepackage{amssymb}
\BibtexOrBiblatex
\electronicVersion
\PrintedOrElectronic
\ifpdf \usepackage[pdftex]{graphicx} \pdfcompresslevel=9
\else \usepackage[dvips]{graphicx} \fi

\usepackage{egweblnk} 
\title{Garment Animation NeRF with Color Editing}

\author[Renke Wang \& Meng Zhang \& Jun Li \& Jian Yang]
{\parbox{\textwidth}{\centering Renke Wang\orcid{0000-0001-6331-3790}, Meng Zhang$^\dagger$\orcid{0000-0003-2384-0697}, Jun Li\orcid{0000-0003-3716-671X}, Jian Yang\thanks{Corresponding author}\orcid{0000-0003-4800-832X}
        }
        \\
{\parbox{\textwidth}{\centering PCA Lab, 
\\
Key Lab of Intelligent Perception and Systems for High-Dimensional Information of Ministry of Education, 
\\
and Jiangsu Key Lab of Image and Video Understanding for Social Security, 
\\
School of Computer Science and Engineering, Nanjing University of Science and Technology
\\
\{wrk226, mengzephyr, junli, csjyang\}@njust.edu.cn
       }
}
}

\begin{document}

% uncomment for using teaser
% \teaser{
%  \includegraphics[width=0.9\linewidth]{eg_new}
%  \centering
%   \caption{New EG Logo}
% \label{fig:teaser}
%}
\teaser{
 \includegraphics[width=\textwidth]{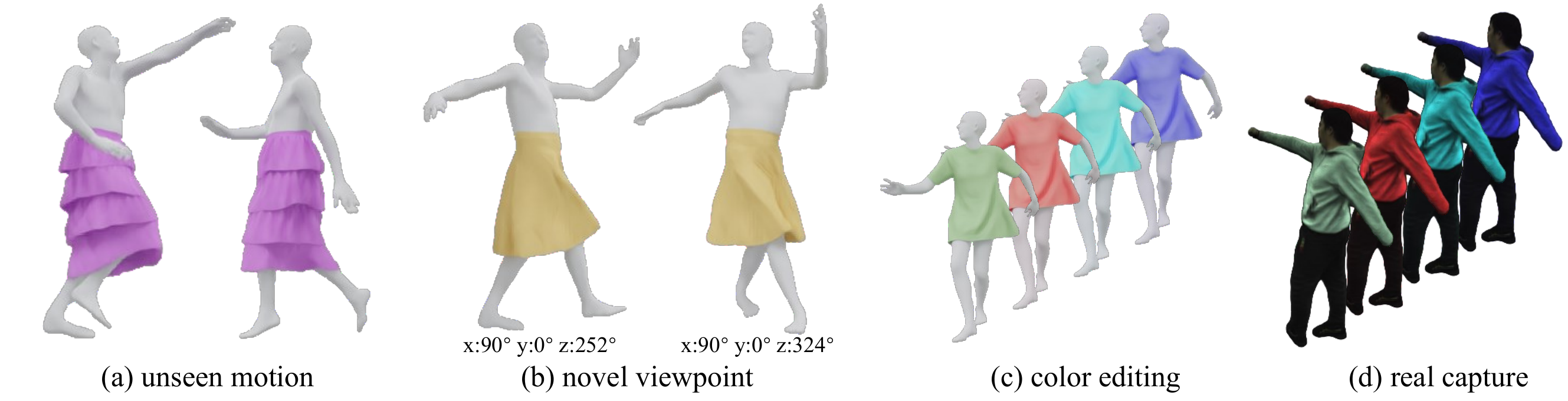}
 \centering
  \caption{
  We introduce a novel \emph{Garment Animation NeRF} that generates character animations directly from body motion sequences, eliminating the need for an explicit garment proxy. Upon training, our network produces garment animations with intricate wrinkle details, ensuring plausible body-and-garment occlusions and maintaining structural consistency across views and frames. We demonstrate the network's generalization capabilities across unseen body motions (a) and camera views (b), while also enabling color editing for garment appearance (c). Notably, our method is applicable to both synthetic (a, b, c) and real-capture (d) garment data.
 % Directly starting from the character's body motion sequence, we present \emph{Garment Animation NeRF} that learns to synthesize the animation of the character dressed in a target garment, with no need for an explicit garment proxy. Once trained, our network synthesizes garment animations with fine wrinkle details while ensuring plausible body-and-garment occlusions and structural consistency across views and frames. We showcase generalization ability of the network across the unseen body motions (a) and unseen camera views (b). Our method allows color editing for the garment appearance (c). Our method is applicable to both synthetic (a, b, c) and real-capture (d) garment data.
  % We present a method of synthesizing garment animation from a given body motion sequence. Our approach does not require garment template, physical simulations or ground truth 3D data. Once trained, our model can synthesize realistic and 3D-consistent loose garment appearances from any viewpoint, based on any unseen motion, and allows for color editing of the target garment appearance. Additionally, our model can be trained using real video sequences to learn the details of the real garments.
  }
    \label{fig:title}
}
\maketitle

%-------------------------------------------------------------------------
\begin{abstract}
Generating high-fidelity garment animations through traditional workflows, from modeling to rendering, is both tedious and expensive. These workflows often require repetitive steps in response to updates in character motion, rendering viewpoint changes, or appearance edits. Although recent neural rendering offers an efficient solution for computationally intensive processes, it struggles with rendering complex garment animations containing fine wrinkle details and realistic garment-and-body occlusions, while maintaining structural consistency across frames and dense view rendering. In this paper, we propose a novel approach to directly synthesize garment animations from body motion sequences without the need for an explicit garment proxy. Our approach infers garment dynamic features from body motion, providing a preliminary overview of garment structure. Simultaneously, we capture detailed features from synthesized reference images of the garment's front and back, generated by a pre-trained image model. These features are then used to construct a neural radiance field that renders the garment animation video. Additionally, our technique enables garment recoloring by decomposing its visual elements. We demonstrate the generalizability of our method across unseen body motions and camera views, ensuring detailed structural consistency. Furthermore, we showcase its applicability to color editing on both real and synthetic garment data. Compared to existing neural rendering techniques, our method exhibits qualitative and quantitative improvements in garment dynamics and wrinkle detail modeling.
Code is available at \url{https://github.com/wrk226/GarmentAnimationNeRF}.

\ccsdesc[300]{Computing methodologies~Rendering}
\ccsdesc[300]{Computing methodologies~Animation}
\ccsdesc[300]{Computing methodologies~Neural networks}

\printccsdesc   
\end{abstract}  

\section{INTRODUCTION}
High-fidelity, detailed garment animation is crucial for enhancing user engagement across various applications such as games, movies and virtual/augmented reality. Modeling, simulating and rendering dynamic garments with realistic folds and wrinkles necessitates a computationally intense processing pipeline, typically operated on expensive professional setups. Any further update to garment appearance, dynamics or viewing camera positions is a tedious task that requires repeating large portions of the pipeline.  

%A revised workflow suggests initially employing coarse garment models to leverage their relatively lower computational complexity, and subsequently, to refine the garment details by learning from multi-view \cite{habermann2021real,xiang2021modeling} or single-view images \cite{icon,econ,pifu,pifuhd}. It can also incorporates data from scans\cite{lahner2018deepwrinkles}, or synthetic garment datasets \cite{zhang2021deep} to enhance the garment wrinkle details. 

Neural rendering represents a significant breakthrough, enabling to learn neural features for controllable image synthesis, including changes in viewpoint and modeling deformations \cite{https://doi.org/10.1111/cgf.14507}.   
The neural rendering methods in garment context \cite{neuralactor, chen2023uv, sanerf, neuralbody}, learns integrating clothing simulation and rendering from multi-view image data. Such techniques enable to synthesize the animation of actors wearing target garments. An intriguing alternative involves the use of learning-based method to translate dynamic garment video synthesise from 2D human cues, such as body pose \cite{aberman2019deep, chan2019everybody, dong2019fw}, body-part segmentation \cite{zhou2019dance}. However, those methods are restricted to tight clothing and don't have good performance on cases of loose and complex garments under dynamic body motions. Dynamic Neural Garments (DNG) \cite{dynamicneuralgarment} employ learnable neural texture for a coarse garment proxy and translate neural descriptor maps to dynamic garment appearance renderings. While demonstrating impressive results with vivid fine wrinkle details, due to a lack of 3D spatial awareness during network inference, DNG struggles to maintain detailed structural consistency across dense views, and requires resolving of the garment-and-body occlusion in the post-processing.

In this work, given a sequence of body motion as input, our work aims at synthesizing dynamic garment animations with fine wrinkle details, without coarse garment proxy predefined \cite{dynamicneuralgarment} or explicitly reconstructed from multi-view images \cite{habermann2021real, xiang2022dressing, xiang2023drivable}, meanwhile, ensuring detailed structural consistency across views. On supervision of multi-view video data, we propose an architecture of neural networks to learn (i) garment dynamic mechanism, (ii) fine wrinkle details, (iii) and structural consistency across view points. 
%\review{for the target garment, enabling generalization ability to various body shapes and body motions}. 
Though trained garment-specific, our network has the generalization ability to synthesize target garment animations for a variety of human body shapes and motion sequences that are reasonable close to the training data.
Moreover, we achieve an artefact-free color editing of the target garment appearance.  

To this end, we first infer a garment dynamic feature map from the geometric and dynamic information of the body, by recording the historical data in the body template texture space. This map provides a preliminary structural overview of the garment's dynamics influenced by the body's movements.
Simultaneously, we encode wrinkle detail feature maps from synthesized garment images using a pre-trained image generative model. To minimize wrinkle detail discrepancies between views caused by the generative model, we carefully select synthesized images from the front and back views to generate detail feature maps with minimal overlap.
In the final rendering stage, we sample the dynamic and detail features according to the radiance point projection on the body's geometry. Additionally, we calculate body-relative geometric features for the radiance points to enhance the 3D spatial awareness with respect to the moving body. Concatenating those features as input, we construct a neural radiance field (NeRF) \cite{nerf} to render garment appearance feature images. Furthermore, we utilize a decoder network to decompose the appearance feature image into several components according to palette base colors to achieve neat segmentation of the target garment. By disentangling the light effect from the pixel color and computing semantic layer-based blending weights, we allow the color editing of garment appearance, maintaining the detail structural realistic.
Consequently, our method efficiently renders plausible garment animations driven by body motions, maintaining detailed structural consistency across dense views. Additionally, it facilitates the recoloring of garment appearances, enhancing the adaptability and utility of the rendered animations. For example, it allows for a quick preview of a garment in various colors.

We evaluate our algorithm on a variety of real and synthetic garments with varying body motions. We showcase that our method generalizes effectively across different body motions and dense view rendering. Additionally, we edit the color of garment appearance, preserving realistic dynamic garment structure. Our method quantitatively and qualitatively surpasses three benchmark techiques: SANeRF \cite{sanerf}, UVVolume\cite{chen2023uv}, and DNG \cite{dynamicneuralgarment}. It delivers convincing garment dynamics, distinctive wrinkle details, accurate body-and-garment occlusions (see Figure \ref{fig:compare} and Table \ref{tab:compare}).

In summary, our key contributions are:
\begin{itemize}
    \item \emph{body-motion-based NeRF} to synthesize the animation of a complex target garment, only taking a body motion as input, with no need of an explicit coarse garment proxy; 
    \item \emph{generalization} over unseen body motions to render garment animations with fine wrinkle details, maintaining structural consistency across views and frames;
    \item \emph{palette-based neural rendering} to allow color editing of target garment appearance free-from artefacts. 
\end{itemize}

\section{RELATED WORK}

\subsection{Modeling garment dynamics}

The employment of physically based simulations has demonstrated effectiveness in achieving authentic depictions of cloth dynamics \cite{choi2005research,liang2019differentiable,narain2012adaptive,nealen2006physically,tang2018cloth,yu2019simulcap}. While yielding highly accurate results, this approach is notably computationally intensive. Although subsequent endeavors have introduced acceleration schemes \cite{li2020p,wu2020safe}, the computational efficiency and stability of these methods remain still largely impacted by the complexity of the garment.

%The use of physically based simulations has proven successful in achieving realistic representations of cloth dynamics \cite{choi2005research,liang2019differentiable,narain2012adaptive,nealen2006physically,tang2018cloth,yu2019simulcap}. It gives very accurate results, but is highly computationally expensive. Subsequent work has introduced some acceleration schemes \cite{li2020p,wu2020safe}, but the computational performance and stability of these methods is still largely influenced by the complexity of the garment. 

To tackle these challenges, data-driven approaches have emerged by involving modeling garment deformations based on pose and body shape \cite{guan2012drape,ma2020learning,patel2020tailornet,santesteban2019learning,li2023animatable}, typically utilizing linear blend skinning. While effective and computationally efficient, they are constrained by their reliance on skinning, which limits their capacity to represent loose garments like long skirts. To handle this, some approaches \cite{santesteban2021self,lin2022learning} propose diffusing skinning weights from the body to the garment's surroundings to capture loose garment dynamics. Alternatively, Pan et. al. \cite{pan2022predicting} introduce virtual bones dedicated to garments to model their dynamics, showing promise in handling loose garments. Unfortunately, these methods often necessitate extensive physical simulations to acquire sufficient 3D data for training. To mitigate this acquirement, physically based constraints as energy losses are leveraged for unsupervised training. For example, Grigorev et. al. \cite{grigorev2023hood} employ hierarchical graph neural networks constructed from garment mesh vertices and edges to transfer physical information, while Bertiche et. al. \cite{bertiche2022neural} directly predict garment deformations based on human motions. These techniques bypass the need for 3D ground truth and have demonstrated impressive performance. However, they face limitations regarding available garment templates and computational resources, and struggle to address self-intersection issues within garments, posing challenges in modeling garments with complex geometries.

In contrast, our model does not necessitate 3D ground truth or garment templates as input, and learns the garment geometry and dynamics directly from multi-view images. This circumvents the manual modeling process and eliminates the requirement for self-intersection detection, significantly augmenting our capability to represent garments with intricate geometries.
%Our model, on the other hand, does not rely on 3D ground truth or garment templates as input, and can learn the corresponding garment geometry and dynamics directly from multi-view images, bypassing the manual modeling process and the need for self-intersection detection. This greatly enhances our ability to model garments with complex geometries.

\subsection{Neural Rendering}
In recent years, neural rendering has demonstrated remarkable success in generating novel views of both static scenes \cite{nerf,wang2021neus,3dgaussiansplatting} and dynamic scenes \cite{pumarola2021d,tretschk2021non,cao2023hexplane,fridovich2023k,luiten2023dynamic}. As a result, researchers have begun exploring the application of neural rendering techniques to model garment dynamics, aiming to provide an integrated solution for simulation and rendering. A key concept in this endeavor is to learn neural features based on 3D human representations, such as skeleton \cite{dynamicneuralgarment, kappel2021high, karthikeyan2024avatarone, noguchi2021neural, liu2023hosnerf, kwon2024deliffas} or human body \cite{neuralbody, neuralactor, sanerf, chen2023uv, gao2023neural,weng2022humannerf}, to control rendering output and accurately reflect body motion and corresponding garment dynamics. For example, NeuralBody \cite{neuralbody} utilizes a vertex-based latent code to establish temporal correspondence and employs a spatially sparse convolution network to convert these codes into a radiance volume, producing high-quality results in novel views. Unfortunately, its rendering quality is only guaranteed for motions seen during training. To address this limitation, researchers have proposed various approaches to establish shared appearance carriers for different poses. Neural Actor \cite{neuralactor} utilizes a canonical space combination with inverse skinning to enable unseen pose generalization. Surface-aligned NeRF \cite{sanerf} constructs an implicit field aligned with the mesh surface to capture surface-dependent appearance, while UV Volume \cite{chen2023uv} decomposes a dynamic human into 3D UV volumes and a 2D texture map, effectively reconstructing consistent appearance under different poses. However, these methods typically model each frame of motion separately, overlooking the continuity of body motion and encountering difficulties in capturing the dynamics of loose garments.

In contrast to previous methods, we incorporate both velocity and normal extracted from human motion as dynamic features in dynamic modeling. This enables us to achieve dynamic-aware and structure-consistent appearance under unseen motion. Additionally, we employ a 2D detail generator to provide wrinkle details for the target garment, enhancing the richness of detail in our results.

\subsection{Image-to-image Translation}
The advent of advanced generative models \cite{goodfellow2014generative, ho2020denoising} has facilitated image translation across domains, leading to the proliferation of high-quality translation techniques \cite{chu2017cyclegan, choi2020stargan,cheng2023general,kang2023scaling}. Early successes like Pix2Pix \cite{isola2017image} demonstrated impressive results by aligning pixel data for paired image translation, effectively transferring both style and content. Expanding on this, some studies have utilized continuous driving signals, such as key points, to achieve realistic video generation \cite{narain2012adaptive, siarohin2019animating, siarohin2019first}. In the realm of human animation and virtual try-on, some approaches focus on using guides such as semantic maps \cite{esser2018variational,fu2022stylegan}, skeletons \cite{aberman2019deep,chan2019everybody,dong2019fw}, or dense correspondences \cite{wang2018video,liu2019liquid,zablotskaia2019dwnet}. These cues aid in aggregating information across frames and generating coherent representations across different movements. However, these methods often struggle to model loose garments due to challenges in establishing pixel-level correspondences. DNG \cite{dynamicneuralgarment} proposed using a coarse garment as a proxy to capture human motion features, enabling realistic garment-driven effects from various viewpoints. Nonetheless, this approach faces difficulties in maintaining detailed structural consistency across different views due to a lack of 3D spatial awareness during inference. Comparatively, our method involves leveraging the detail map generated in image-to-image schemes and aligning it with a pose-aware neural radiance field, resulting in garment dynamics that are not only detailed but also structurally consistent.

\subsection{Garment authoring}
With the advancement of digital fashion, several algorithms have emerged enabling intuitive garment editing for non-professionals. These include generating 3D models from photos \cite{zhou2013garment} or sketches \cite{de2015secondskin,wang2018learning}, resizing garments for different body shapes \cite{meng2012flexible}, designing textures from wallpapers \cite{wolff2019wallpaper}, and crafting various styles of tight-fitting clothes \cite{kwok2015styling}. Recent developments also facilitate direct adjustments on 3D garments \cite{pietroni2022computational} without simulations. Our research extends these technologies by focusing on garment animation modeling and allowing color editing without artifacts.

% Recently, with advances in digital fashion design, a number of algorithms have been proposed to allow intuitive editing of garments, enabling non-professionals to easily create and modify digital garments. For example, generating 3D garment models directly from a single photo \cite{zhou2013garment} or sketch \cite{de2015secondskin,wang2018learning}, automatically scaling garments to fit human bodies \cite{meng2012flexible}, adding appropriate textures pattern based on given wallpaper \cite{wolff2019wallpaper}, or even generating different styles of tight-fitting garments \cite{kwok2015styling}. To provide more intuitive editing capabilities and lower the barriers to garment design for non-professionals, some models have been developed that allow adjustments directly on 3D garments \cite{pietroni2022computational} or 2D patterns \cite{umetani2011sensitive} without the need for physical simulations. These approaches are all based on static garments, whereas our work focuses primarily on modeling dynamics, and thus complements these efforts by learning dynamic effects for these garments. In addition, our model allows color editing of the target garment appearance free of artifacts.

\section{OVERVIEW}
%\review{...camera random posed circle around}
Given a set of multi-view RGB videos captured circle around an motion sequence of a character dressed in a complex or loose target garment, 
%Given a multi-view RGB video of a character dressed in a complex or loose garment and the corresponding body motion sequence during the training period, 
our method is designed to synthesize garment animation in image space when given a new (unseen) human motion sequence and arbitrary view points. We render garment animations with high-quality fine wrinkles, ensuring structural consistency across dense-view rendering. Moreover, our technique allows for color editing of the target garment.

\begin{figure*}[t]
\centering
\includegraphics[width=1.\linewidth]{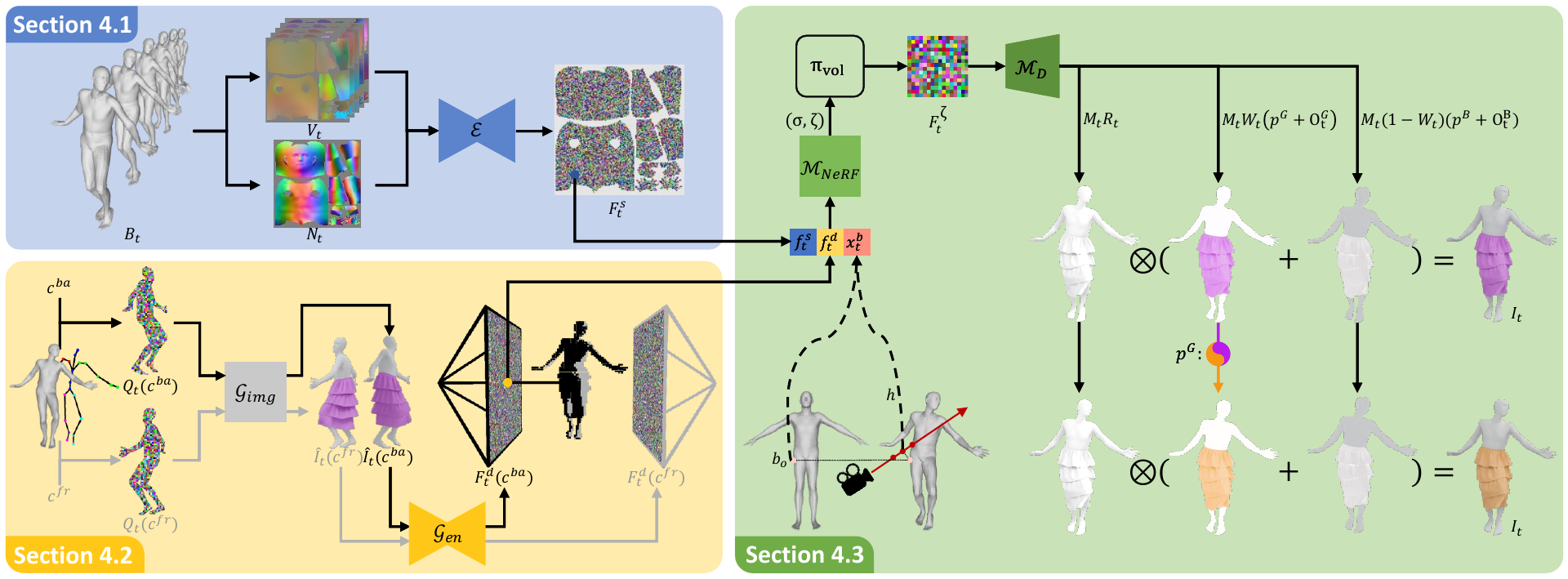}
\vskip -0.1in
\caption{\textbf{The architecture of \emph{Garment Animation NeRF}.} 
Given a sequence of character's body motion, we construct a neural radiance field to render animation of the character dressed in the target garment. 
We first employ a dynamic feature encoder $\mathcal{E}$ to infer garment dynamic feature map $F_t^s$ from the information textures $V_t$ and $N_t$ of the body motion $B_t$.
Simultaneously, taking body neural texture images $Q_t$ of the body at front $c^{fr}$ and back views $c^{ba}$, we use a pre-trained image generator $\mathcal{G}_{img}$ to predict reference images $\hat{I}_t(c^{fr})$ and $\hat{I}_t(c^{ba})$. Subsequently, we use the detail feature encoder $\mathcal{G}_{en}$ to generate the detail feature maps $F_t^d(c^{fr})$ and $F_t^d(c^{ba})$. 
Then, we obtain body-aware geometric information $x_t^b$ by calculating the distance $h$ between sampling points and the body surface, and finding $b_o$ on the canonical body shape.
Finally, we utilize a NeRF network $\mathcal{M}_{NeRF}$ to render garment appearance feature image $F_t^{\zeta}$. To enable color editing, we introduce a network $\mathcal{M}_D$ to decompose the garment appearance into a front mask $M_t$, a color offset map $O_t$, a radiance map $R_t$ and a blending weight map $W_t$. By linearly recombining those visual elements, we synthesize the final frame image $I_t$. Except for the generator $\mathcal{G}_{img}$, we jointly train the networks of $\mathcal{E}$, $\mathcal{G}_{en}$, $\mathcal{M}_{NeRF}$ and $\mathcal{M}_{D}$ in an end-to-end manner.
% Given a sequence of body motions over k frames, we align features by sampling points along the camera ray around the body. 
% Then we 1. extract detail feature $f_t^d$ from the gray-scale images $\hat{I}_t$ generated by the image generator $\mathcal{G}_{img}$. 2. Use the UV map to record the surface normal and velocity and encode with dynamic feature encoder $\mathcal{E}$, thereby extracting the dynamic structural feature $f_t^s$. 3. obtain the body aware information $x_t^b$, which include the distance $h$ between the sample point and the body mesh and the projection point $b_t$ on the T-pose mesh. The extracted features $f_t^d$, $f_t^s$ and $x_t^b$ are then concatenated and input into $\mathcal{M}_{NeRF}$ to construct a feature radiance field. The feature radiance field is then aggregated by volume rendering and fed into $\mathcal{M}_D$, which allows for the prediction of the current scene's front mask ${M}_t$, color offset map $O_t$, radiance map $R_t$ and blending weight map $W_t$. 
% These elements are then integrated cohesively using Eq. \ref{eq:palette_combine} to synthesize the final image.
}\label{fig:pipeline}
\vskip -0.15in
\end{figure*}

We propose an architecture of  \emph{Garment Animation NeRF}, that composes of three components: the \emph{dynamic feature encoder} $\mathcal{E}$, the \emph{detail feature generator} $\mathcal{G}$, and the \emph{rendering network} $\mathcal{M}$. We show the network architecture in Figure \ref{fig:pipeline}.
Initially, we learn the dynamic features of the garment influenced by the human body motion, by encoding the texture of body motion features using an encoder network $\mathcal{E}$. Simultaneously, to generate the garment dynamic wrinkle features, we utilize a generative model $\mathcal{G}$ to create neural images of garment detail features, taking the front and back views of the body motion as input. In the final stage, we sample the garment dynamic and detail features based on the radiance point projection and compute the 3D body-aware geometric information, which are then concatenated as inputs for the NeRF $\mathcal{M}_{NeRF}$ to render the feature image that depicts the garment appearance. Furthermore, we develop an innovative network $\mathcal{M}_{D}$, that integrates with $\mathcal{M}_{NeRF}$, for palette-based decomposition, enabling the color editing of the target garment. This is achieved by decomposing the color elements from the appearance features, and then recombining them through a linear combination, thus generating the final target garment image.

% Our pipeline consists of three main components: \emph{3D feature extractor}, which extracts a UV-based body motion feature map used to describe the dynamics of the loose garment influenced by human body motion; \emph{2D Feature Extractor}, uses a 2D generative model to generate a pair of reference images and obtain detailed cues for detail enhancement. 3. The renderer uses the previously extracted features to build a Pose-Invariant Garment NeRF, then decomposes the visual elements through palette-based 2D neural rendering, allowing the target garment to be recolored.

%-------------------------------------------------
%-------------------------------------------------
%-------------------------------------------------
% \begin{figure*}[h]
% \centering
% \includegraphics[width=1.\linewidth]{images/symbol-table.png}
% \vskip -0.1in
% \caption{Symbol table.}\label{fig:pipeline}
% \vskip -0.15in
% \end{figure*}
%-------------------------------------------------

\section{ALGORITHM}

\subsection{Dynamic Feature Encoder} \label{sec: Dynamic Feature Encoder}
Our method first infers garment dynamic structural features from the input of desired 3D body motion. Given a body motion $B_t$ at current time $t$, we employ UV texture space to capture body geometric and dynamic information, thereby obtaining the body shape normal map ${N}_t\in \mathbb{R}^{128\times128\times3}$ and the velocity map ${V}_t\in \mathbb{R}^{128\times128\times3k}$, where $k$ denotes the number of historical frames that dynamic information ${V}_t$ contains. Utilizing the body information texture $\{{N}_t, {V}_t\}$, we employ the dynamic feature encoder $\mathcal{E}$, comprised of a series of 2D convolution layers, to encode the dynamic features of garments. We compute the implicit dynamic feature map {${F}_t^s \in \mathbb{R}^{128\times128\times8}$} by transforming the body texture features as:
\begin{equation}
{F}^s_t=\mathcal{E}({N}_t,{V}_t).
\end{equation}
The dynamic feature map ${F}^s_t$ of the garment provides a preliminary structural overview of the garment dynamics and the interactions between body and garment. This aids in ensuring garment structural consistency across views and in generating plausible occlusions between the body and the garment. 
Please note that, unlike methods that map dynamic features to the garment UV space \cite{dynamicneuralgarment, habermann2021real, xiang2022dressing, xiang2023drivable}, we infer garment dynamic features soly from body dynamics, eliminating the need to project features onto any predefined garment proxy space.
% Given a 3D body motion as input, we first use UV space to record its surface normal and velocity, thereby obtaining the surface normal map $\mathbf{N}\in \mathbb{R}^{128\times128\times3}$ and the velocity map $\mathbf{V}\in \mathbb{R}^{128\times128\times3t}$, where $t$ represents the number of historical frames that dynamic information $\mathbf{V}$ contains.
% The 2D representation of surface normal and velocity allows us to further use 2D CNNs to capture the relationships between adjacent body parts through local convolutional operations. Since each pixel on the UV map corresponds to a specific spatial point on the body surface, we can concatenate the previously obtained $N$ and $V$, then extract the 3D feature map $\mathbf{F}_{3D}\in \mathbb{R}^{128\times128\times128}$ for implicit garment dynamics modeling with the 3D feature encoder $\mathcal{E}_{3D}$:
% \begin{equation}
% \mathbf{F}_{3D}=\mathcal{E}_{3D}(\mathbf{N},\mathbf{V}), \mathbf{F}_{3D}\in\mathbb{R}^{128\times128\times128},
% \end{equation}
%We assign the extracted features to the spatial sample points $\mathbf{x}$ closest to them in the neural radiance field.

\subsection{Detail Feature Generator} \label{sec: detail feature generator}
Inspired by DNG \cite{dynamicneuralgarment}, we employ a generative model $\mathcal{G}$, to synthesize the detail features of garment wrinkle details, with the desired body motion serving as input. In contrast to DNG, which uses a coarse garment proxy to guide the dynamic neural rendering, our approach directly apply the learnable neural textures to the body geometry and renders neural images $Q_t$ to serve as inputs of the detail feature generator $\mathcal{G}$.  

Our detail Feature generator, $\mathcal{G}$, is composed of two parts: the image generator $\mathcal{G}_{img}$ and the detail feature encoder $\mathcal{G}_{en}$. Following the methodology of DNG, we pre-train $\mathcal{G}_{img}$ with learnable neural texture images that capture body dynamics, to predict reference images $\hat{I}_t$ of garment dynamics, sized at $512\times 512$. The training is supervised by ground truth images of the garment. The pre-trained image-based generative model $\mathcal{G}_{img}$ provides essential hints for rendering garment appearance details.

Taking body neural texture images of the body $Q_t(c)$ at a specific camera pose $c$ as input, we use the image generator $\mathcal{G}_{img}$ to predict the reference image $\hat{I}_t(c)$. Then we use the detail feature encoder $\mathcal{G}_{en}$ to generate the detail feature maps ${F}_t^d(c) \in \mathbb{R}^{512 \times 512 \times 128}$ as: 
\begin{equation}
{F}_t^d(c)= \mathcal{G}_{en}[\hat{I}_t(c)]=\mathcal{G}_{en}[\mathcal{G}_{img}[Q_t(c)]]. 
\end{equation}

To mitigate the negative impacts of detail inconsistencies between views, as observed in \cite{dynamicneuralgarment}, we employ the image generator $\mathcal{G}_{img}$ to render only the front and back views to compute garment detail feature maps, ${F}_t^d(c^{fr})$ and ${F}_t^d(c^{ba})$. This ensures that the rendered details features maps are free from detail structural conflicts. 
Please note that, as trained with multi-view videos captured circle around the character wearing the target garment, the image generator $\mathcal{G}_{img}$ has the generalization ability to render the garment animation at front and back viewpoints, even if they are located outside the training camera viewpoints.

\subsection{Rendering Network} \label{sec: Rendering Network}
Our rendering network $\mathcal{M}$ is mainly composed of a NeRF network $\mathcal{M}_{NeRF}$ to render garment appearance feature image, and a network $\mathcal{M}_D$ for palette-based decomposition to allow color editing of the target garment.

\textbf{Appearance feature image.} 
Our NeRF network $\mathcal{M}_{NeRF}$ constructs a temporal-and-spatial aware dynamic implicit field, adaptive to body dynamics. It aims to learn three key factors: (i) the garment's dynamic structure driven by body motion;(ii)wrinkle details enhancement ensuring temporal coherence; and(iii) occlusions between garment layers or between the garment and the body. To achieve, we first construct 3D garment dynamic feature distribution with features sampled from the acquired dynamic structural feature map $F_t^s$. Next, we project generated detail features back to 3D space to concatenate them with the garment features and body-aware spatial information. Finally, while constructing the 3D dynamic garment appearance field, we compute a density field around the moving body to indicate the occlusion between the garment and the body, or within the garment's structural layers.

With a 3D point position $x$ sampled along the camera ray towards a specified image pixel, we begin by projecting $x$ to the posed body geometry $B_t$ to obtain the projection point $b_t$. We then calculate the distance $h=|x-b_t|$ between $x$ and $b_t$. To enable pose-invariant body-relative sampling, we transform $b_t$ back to the canonical pose, yielding the position $b_o$ on the canonical body shape $B_o$. This allows us to represent the body-aware geometric information as $x_t^b := [b_o, h]$. Utilizing the corresponding body UV space coordinate of $b_o$, we sample an implicit garment dynamic feature $f_t^s$ from the acquired dynamic structural feature map $F_t^s$. Simultaneously, we sample the detail feature $f_t^d$ from the detail feature image rendering, either at the front $F_t^d(c^{fr})$ or back view $F_t^d(c^{fr})$, depending on whether the projection point $b_t$ is visible at the camera pose $c^{fr}$ or $c^{ba}$. 

Given inputs of {dynamic feature sampling $f_t^s$, detail feature sampling $f_t^d$, and body-aware geometric information $x_t^b$}, our NeRF network, $\mathcal{M}_{NeRF}$, computes the density $\sigma \in \mathbb{R}^1$ and garment appearance feature $\zeta \in \mathbb{R}^{128}$ at the 3D point position $x$. This computation is expressed as follows:
\begin{equation}
\sigma, \zeta=\mathcal{M}_{NeRF}[f_t^s, f_t^d, x_t^b], 
\end{equation}
where $\mathcal{M}_{NeRF}$ consists of multiple layers of perceptrons. Subsequently, we employ volume rendering $\pi_{vol}$ to generate the garment appearance feature image $F_t^{\zeta} \in \mathbb{R}^{128\times128\times128}$ as:
\begin{equation}
F_t^{\zeta}=\pi_{vol}(\zeta,\sigma).
\end{equation}
The volume rendering $\pi_{vol}$ adheres to the integration methodology detailed in \cite{mildenhall2020nerf}, with the modification of substituting the color element with the appearance feature $\zeta$. 

\textbf{Palette-based decomposition.} 
Inspired by \cite{palettenerf}, we propose a 2D convolution neural network $\mathcal{M}_D$ to decompose the appearance feature image $F_t^{\zeta}$ into multiple visual elements. These elements are then linearly blended using a predicted weight map to generate the final target garment image. 
%\review{In contrast to \cite{palettenerf}, we do not perform color decomposition at each sampling point. Instead, we use $\mathcal{M}_D$ to directly decompose the feature map. This approach allows each pixel to perceive the information of adjacent pixels through convolution operations while predicting each visual element, which helps to further optimize the volume rendering results.}

To extract the palette-based colors of interest, we statistically compute the mean colors, $\mu^G$ and $\mu^B$, for both the regions of target garment $G$ and body $B$. We initialize the palette base color vector $p:=[p^G, p^B]$ with $p^*=[\mu^G, \mu^B]$, that $p \in \mathbb{R}^{6}$. With the acquired appearance feature map $F_t^{\zeta}$ as input, the network $\mathcal{M}_D$ up-samples the feature image from the size of $128$ to $512$, and meanwhile, decouples multiple visual element maps, specifically, the color offset map $O_t$, radiance map $R_t$, blending weight map $W_t$. This computation is expressed as follows:
\begin{equation}
O_t, R_t, W_t, {M}_t= \mathcal{M}_D[F_t^\zeta].
\label{eq:palette_combine}
\end{equation}
To focus on rendering the moving body dressed with the target garment, $\mathcal{M}_D$ also predicts a mask $M_t$ of the region of both garment and body. 

For each image pixel $i$ in the decoupled map, its color offset element, $o_t := [o_t^G, o_t^B], o_t \in \mathbb{R}^6$ in the map $O_t$, captures its offset with respect to the palette-based color vector $p$. The radiance element, $r_t \in \mathbb{R}^3$ in the map $R_t$, represents the intensity of light, while its weight element, $w_t \in \mathbb{R}^1$ in the map $W_t$, determines the likelihood that the pixel $i$ belongs to the garment region. Additionally, $m_t \in \mathbb{R}^1$ in the mask $M_t$, indicates whether pixel $i$ falls within the rendering region of 
both the garment and body. Consequently, the color $c(i)$ of pixel $i$ is computed as:
\begin{equation}
    c(i) = m_t \otimes r_t \otimes [w_t \otimes (p^G+o_t^G) + (1-w_t) \otimes (p^B + o_t^B)],
\end{equation}
where $\otimes$ denotes element-wise vector multiplication. This formulation allows us to modify the color of the target garment by adjusting the palette-based vector $p^G$ to any desired color $\hat{p}^G$, while keeping the other visual elements unchanged.

\subsection{Loss Function}
Except for the generator $\mathcal{G}_{img}$, which is pre-trained following the instruction in \cite{dynamicneuralgarment} supervised by ground truth images as discussed in Section \ref{sec: detail feature generator}, we jointly train the dynamic feature encoder $\mathcal{E}$, the detail feature encoder $\mathcal{G}_{en}$, and the rendering blocks of $\mathcal{M}_{NeRF}$ and $\mathcal{M}_{D}$, in an end-to-end manner.

To ensure the image rendering quality, we employ L1 loss with respect to the colors and masks in the region of interest. We define a loss function $L_{img}$ as follows:
\begin{equation}
    L_{img} = \|I_t-I^*_t\|_1 + \|M_t-M^*_t\|_1.\,
\end{equation}
where, $I_t$ denotes an image synthesized by our network with $I^*_t$ as its ground truth, and $M_t$ denotes the predicted mask with $M^*_t$ as its ground truth. Furthermore, we consider the similarity of multi-layer features $VGG^i[I_t]$ of the pre-trained network VGG and represent the loss function as:
\begin{equation}
    L_{vgg}= \sum_i \|VGG^i[I_t]-VGG^i[I^*_t]\|_1.
\end{equation}
For color editing of the target garment, we adapt the sparsity loss $L_{sp}$ and the color offset loss $L_{off}$ to achieve correct segmentation of the target garment. The sparsity loss $L_{sp}$ is applied on the predicted blending weight map $W_t$, and defined as:
\begin{equation}
    L_{sp} = \|\frac{1}{W_t^2 + (1-W_t)^2}-1\|_1.
\end{equation}
And the offset loss $L_{off}$ is applied on the predicted color offset map $O_t$, and defined as:
\begin{equation}
    L_{off} = \|O_t\|_2^2.
\end{equation}
Following \cite{palettenerf}, the sparsity loss $L_{sp}$ and the offset loss $L_{off}$ act as adversarial roles to achieve a neat segmentation of the target garment region, with the optimization of the palette base color vector $p$ by applying the loss $L_{p}$ defined as:
\begin{equation}
    L_{p} = \|p-p^*\|_2^2,
\end{equation}
where $p^*$ is the initialization of the base vector introduced in Section \ref{sec: Rendering Network}. 
The final loss function we use to train our network is a weighted sum of the terms:
\begin{equation}
    L = \lambda_1 L_{img} + \lambda_2 L_{vgg} + \lambda_3 L_{sp} + \lambda_4 L_{off} +\lambda_5 L_{p},
\end{equation}
where we set $\lambda_1=1, \lambda_2=0.1, \lambda_3=0.0002, \lambda_4=0.03, \lambda_5=0.001$ in our experiments. $\lambda_1$ and $\lambda_2$ were chosen to balance the weights of image and VGG features. $\lambda_3$, $\lambda_4$. $\lambda_5$ were set following \cite{palettenerf}.

\section{RESULTS AND EXPERIMENTS}
In this section, we demonstrate the effectiveness of our approach in various scenarios. We also compare the quality and quantity results of our method against other benchmark solutions. We will first show how our method generalizes to novel camera views and unseen motions. Additionally, we will present the color editing capabilities of our method.

\subsection{Data Generation}
To train our model, we establish a synthetic dataset. We use SMPL \cite{SMPL} model as our body template, and then obtain a motion sequence of 800 frames from Mixamo (https://www.mixamo.com/) for training. Next, we design three sets of garment (t-shirt, skirt, and multi-layer skirt) using Marvelous Designer, and perform physical simulations of clothing with the body motion sequence to obtain the ground truth garment dynamics. Subsequently, we select 16 fixed camera positions on a circle around the body shape to render videos of the moving body dressed in our target garments. Under each view, we generate the ground truth animation $I^*_t$ and corresponding ground truth front mask $M^*_t$.

\subsection{Implementation details}
\textbf{Architecture.} Our dynamic feature encoder $\mathcal{E}$ takes the concatenation of ${V}_t$ and ${N}_t$ as input. It initially employs a convolution layer to map the recorded human body information from $\mathbb{R}^{128\times128\times3(k+1)}$ to $\mathbb{R}^{128\times128\times32}$, where we set $k=2$. Then it's followed by four convolutional layers that gradually downsample the resolution of the feature map from $128\times128$ to $8\times8$ and increase the feature dimensions to 64, 128, 256, 512, respectively. Subsequently, we utilize four convolution upsampling layers to gradually upsample the resolution of feature map back to $128\times128$ and decrease the feature dimensions to 256, 128, 64, 32. At each stage, the output is concatenated with the corresponding downsampled result in a residual-like connection before being passed through the convolution layer. Thus, the encoder network $\mathcal{E}$ computes the dynamic structural feature map $F_t^s$ sized at ${128\times128\times8}$. We apply instance normalization and leakyReLU for all convolution layers except the final one, which uses a tanh activation function.

The detail feature encoder $\mathcal{G}_{en}$ employs a similar autoencoder architecture with residual connections as $\mathcal{E}$. Taking a reference image $I_t(c^{fr})$ or $I_t(c^{ba})$ as input, it first downsamples the resolution from $512\times512$ to $32\times32$ and increases the feature dimension from 3 to 512. Then, it upsamples the feature map gradually back to resolution of  $512\times512$ with 4 convolution layers, and decreases the feature dimension to 256, 128, 64, 32, respectively. An additional convolution layer is used to produce the final reference feature map $F_t^d$ with size of ${512\times512\times8}$. 

$\mathcal{M}_{NeRF}$ comprises six linear layers, each with a latent dimension of 256 and ReLU activation. It takes the concatenation of $x_t^b, f_t^d, f_t^s$ as input with shape $\mathbb{R}^{4+8+8}$. We employ 2 additional linear layers to respectively project the output into $\mathbf{\sigma\in \mathbb{R}^{1}}$ and $\mathbf{\zeta\in \mathbb{R}^{128}}$. We compute the appearance feature map $F_t^{\zeta}$ with accumulation processing along pixel rays, following the methodology of volume rendering.

With the appearance feature map $F_t^{\zeta}\in \mathbb{R}^{128\times128\times128}$ as input,  the network  $\mathcal{M}_D$ upsamples the resolution from 128 to 512
with 2 convolution layers, and encodes the feature dimension from 128 to 12. Finally, we split the output feature map into a color offset map $O_t\in\mathbb{R}^{512\times512\times6}$, a radiance map $R_t\in\mathbb{R}^{512\times512\times3}$, a blending weight map $W_t\in\mathbb{R}^{512\times512\times2}$ and a front mask $M_t\in\mathbb{R}^{512\times512\times1}$.

\textbf{Training.} We start by training the image generator $\mathcal{G}_{img}$ according to the approach outlined in DNG\cite{dynamicneuralgarment}, which is subsequently frozen to facilitate the training of the other modules. Except for $\mathcal{G}_{img}$, we jointly train the networks of $\mathcal{E}$, $\mathcal{G}_{en}$, $\mathcal{M}_{NeRF}$ and $\mathcal{M}_{D}$ in an end-to-end manner. We set the training batch size to be 1 and utilize an Adam optimizer with a learning rate exponentially decaying from \(5 \times 10^{-4}\) to \(5 \times 10^{-5}\). With a single RTX 3090 GPU, our network converges after 200k iterations, taking around 32 hours for training.

% The remain part contain two stages: In the first stage, we directly extract detail feature $f_t^d$ from ground truth image. This step ensures the model efficiently converges, translating the 2D detail feature into the results. In the second stage, we switch to using images generated by $\mathcal{G}$ for reference, enhancing the model's ability to extract detail from inaccurate images and improving generalization to unseen motion. The initial training stage is completed over 160k iterations, followed by 40k iterations in the latter stage, with both stages employing a batch size of one and utilizing the Adam optimizer. the learning rate is set to decay exponentially from \(5 \times 10^{-4}\) to \(5 \times 10^{-5}\). The training process spans approximately 32 hours utilizing a single RTX 3090 GPU.
%-------------------------------------------------
\begin{figure}[t]
\centering
\includegraphics[width=1.\linewidth]{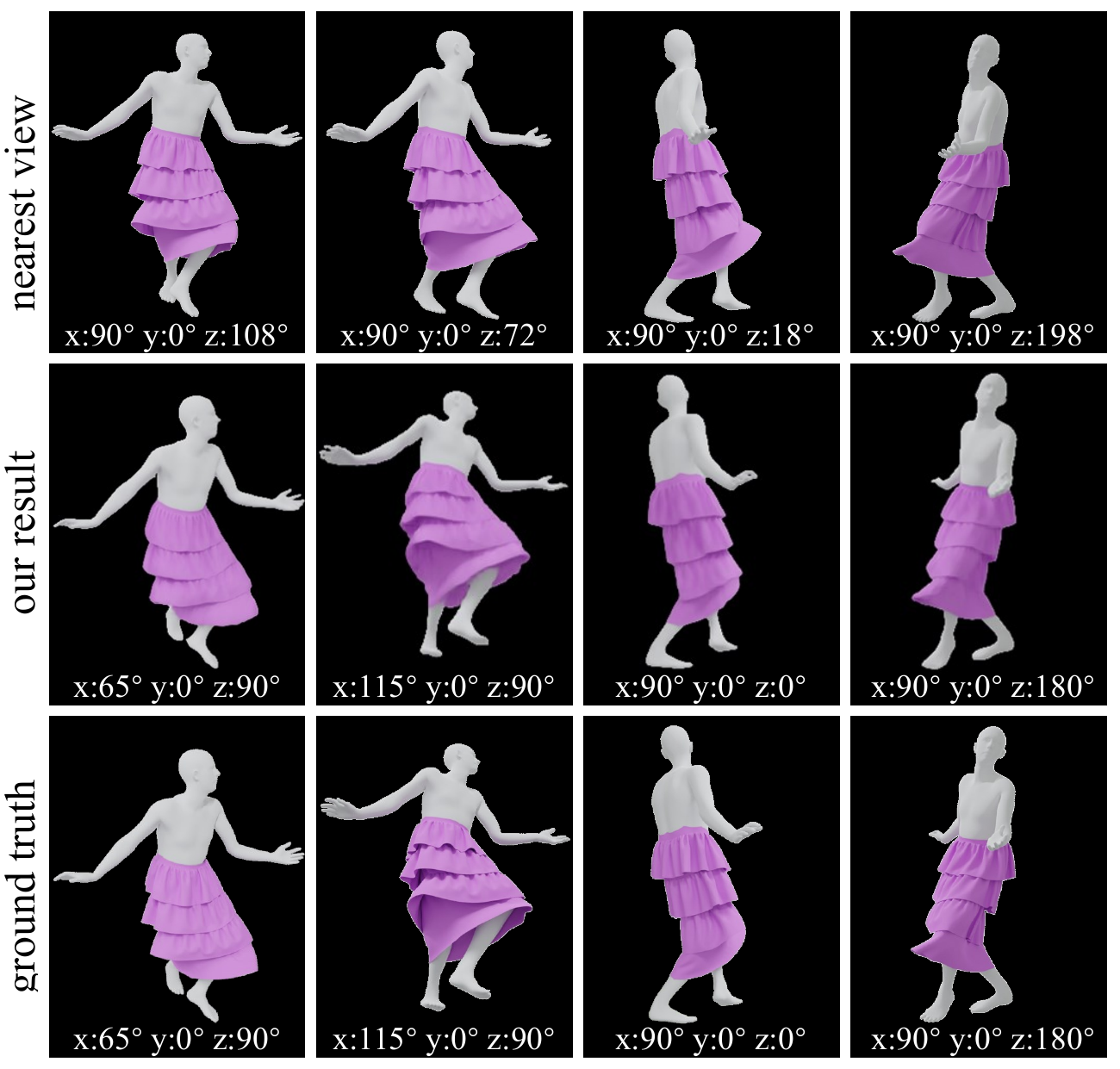}
\vskip -0.1in
\caption{\textbf{Unseen view.} Our model can generate a structural consistent garment appearance from arbitrary viewpoint.}\label{fig:unseen_view}
\vskip -0.15in
\end{figure}
%-------------------------------------------------
\subsection{Results and evaluation}
We now evaluate the generalization ability of our method. Here we show sampled frames and camera views. To better evaluate the visual quality of our garment animation rendering, please refer to the supplementary video.

\textbf{Unseen views.} 
For the motion sequence seen during training, we demonstrate the generalization of our model across unseen camera view points. In Figure \ref{fig:unseen_view}, for each test view, we show the training samples with the nearest viewpoint and the ground truth rendering. As can be seen in the supplemental video and Table \ref{tab:compare}, compared to the nearest training samples, our method generates garment animations in higher performance with respect to perceptual metrics.
% For the motion sequence seen during training, we demonstrate the visual effects of our model from unseen views, as shown in Fig. \ref{fig:unseen_view} and Fig. \ref{fig:compare}. Although this is not the primary focus of our model, it still surpasses the comparative methods in all metrics. This proves that the fusion of dynamic and detail information benefits the novel view synthesis as well.
%-------------------------------------------------

%-------------------------------------------------
\begin{figure}[th]
\centering
\includegraphics[width=0.96\linewidth]{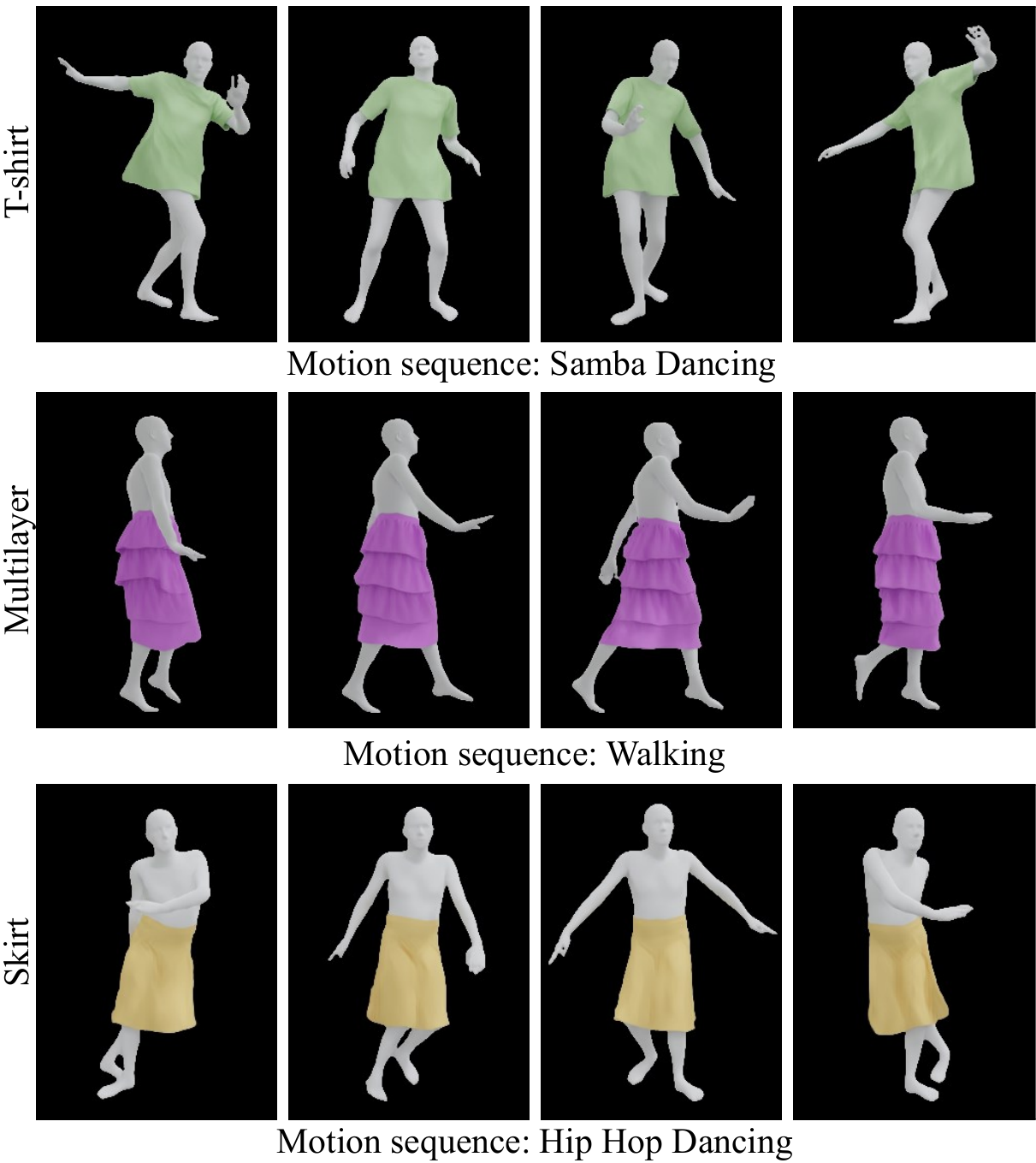}
\vskip -0.1in
\caption{\textbf{Unseen motion.} We test our model on various garments using several motion sequences that were not seen in the training process.}\label{fig:unseen_pose}
\vskip -0.25in
\end{figure}
%-------------------------------------------------

%-------------------------------------------------
\begin{figure}[t]
\centering
\includegraphics[width=.9\linewidth]{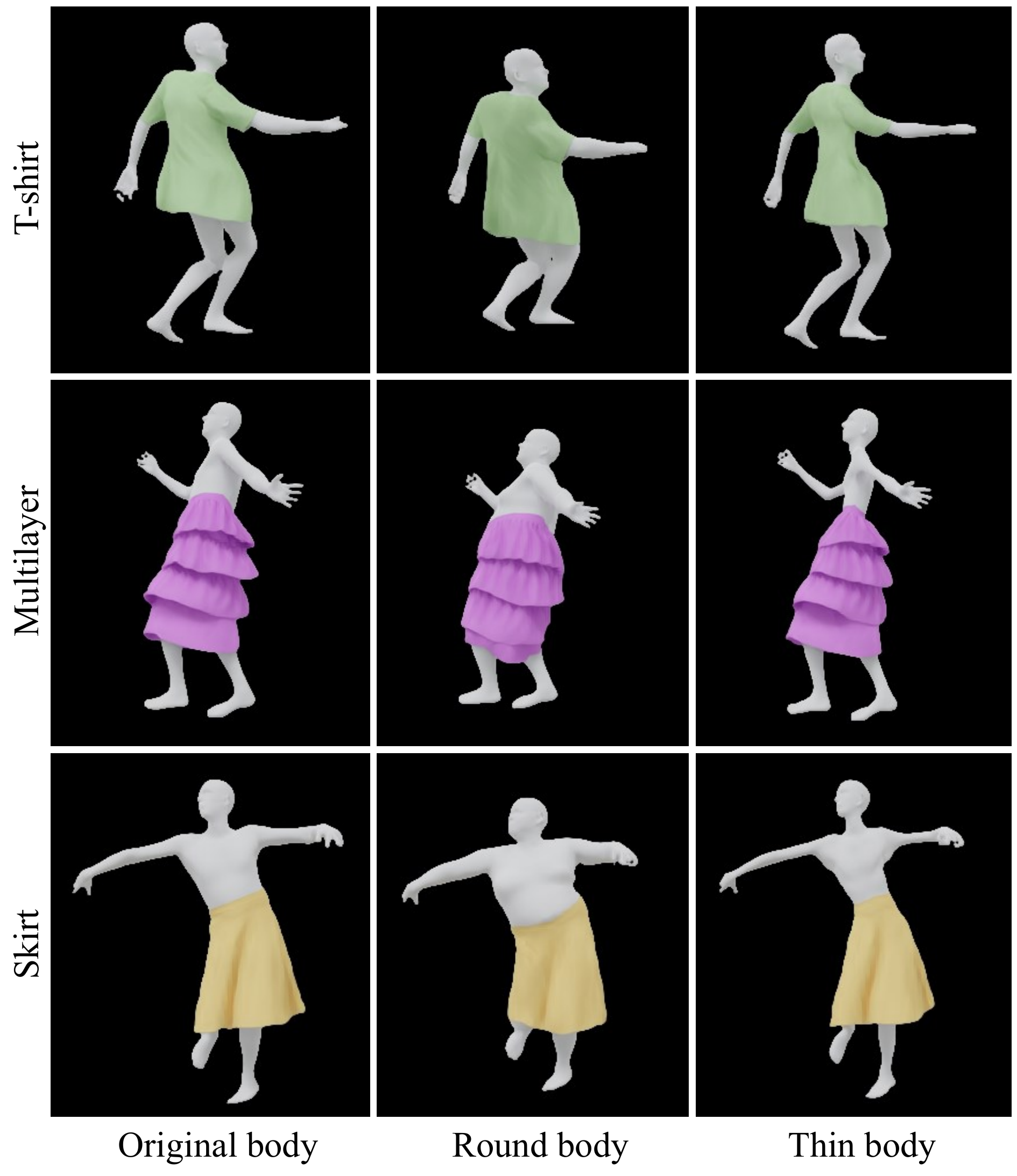}
\vskip -0.1in
\caption{\textbf{Unseen body shapes.} 
Without fine-tuning, our method can synthesize target garment animation that fit with the body shapes (round and thin) different from the original training data body shape.}\label{fig:body_shape}
\vskip -0.1in
\end{figure}
%-------------------------------------------------

%-------------------------------------------------
\begin{figure}[t]
\centering
\includegraphics[width=1.\linewidth]{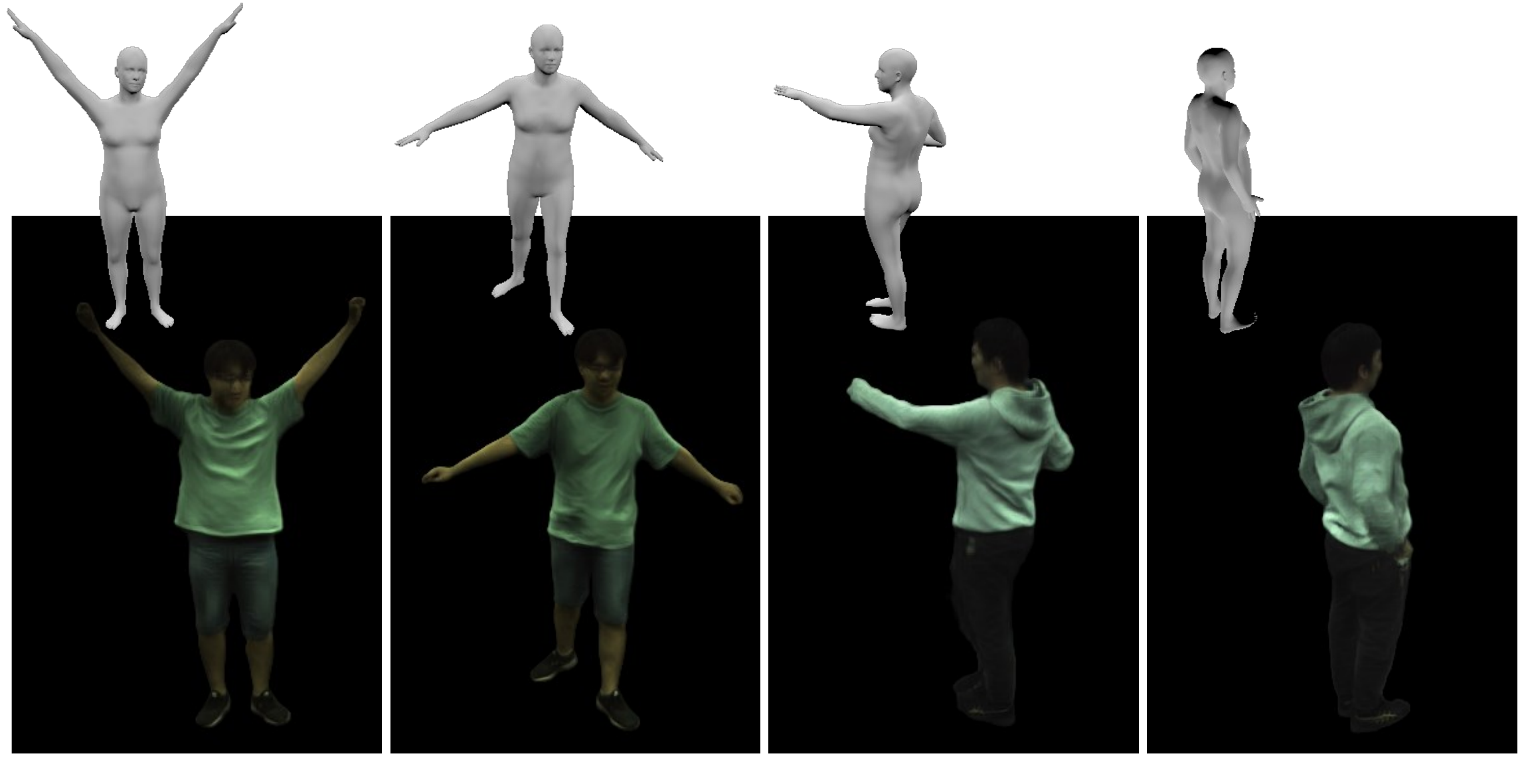}
\vskip -0.1in
\caption{\textbf{Real-captured data.} Our model can be trained with real-captured data and generalize to the body motion unseen during the training.}\label{fig:real_example}
\vskip -0.2in
\end{figure}
%-------------------------------------------------
\textbf{Unseen motion.} Next, we evaluate our model's performance on motion sequences not seen during training. As illustrated in Figure \ref{fig:unseen_pose}, we conducted tests on the t-shirt, multilayer skirt, and long skirt across different motion sequences, including samba dancing, walking, and hip hop dancing. The results demonstrate that our model exhibits good generalization capabilities to unseen motion sequences. It realistically simulates dynamic effects while preserving garments' topological structure and local folds.

\textbf{Unseen body shapes.}
We evaluate the generalization ability to unseen body shapes. 
DNG \cite{dynamicneuralgarment} requires fine-tuning to fit unseen body shapes, as it learns a mapping from the neural texture of the body-independent coarse garment proxy to the target garment animation. In contrast, our method can synthesize garment animations for various body shapes without network fine-tuning, even trained with a fixed body shape. This is because our method synthesizes garment animations based on the distribution of dynamic features $f_t^s$, detail features $f_t^d$ and body-aware information $x^b_t$, which are adaptive and closely related to the body shape.
Figure \ref{fig:body_shape} demonstrates rendering results of three target garment types tested on two body shapes (round and thin) different from the original body shape of training data. 
%We also evaluate our method with different body shapes, as shown in Figure \ref{fig:body_shape}. We first train our model using multiview RGB videos with original body shape. After training our model on multiview RGB videos with the original body shape, we change the input to different, unseen body shapes.Benefiting from the body-aware geometric information $x^b_t$, which provides the garment's relative position to the body surface, our model naturally fits the garment to these new shapes without fine-tuning, demonstrating its accurate learning of 3D interactions between body and garment.

%-------------------------------------------------
\begin{figure*}[t]
\centering
\includegraphics[width=1.\linewidth]{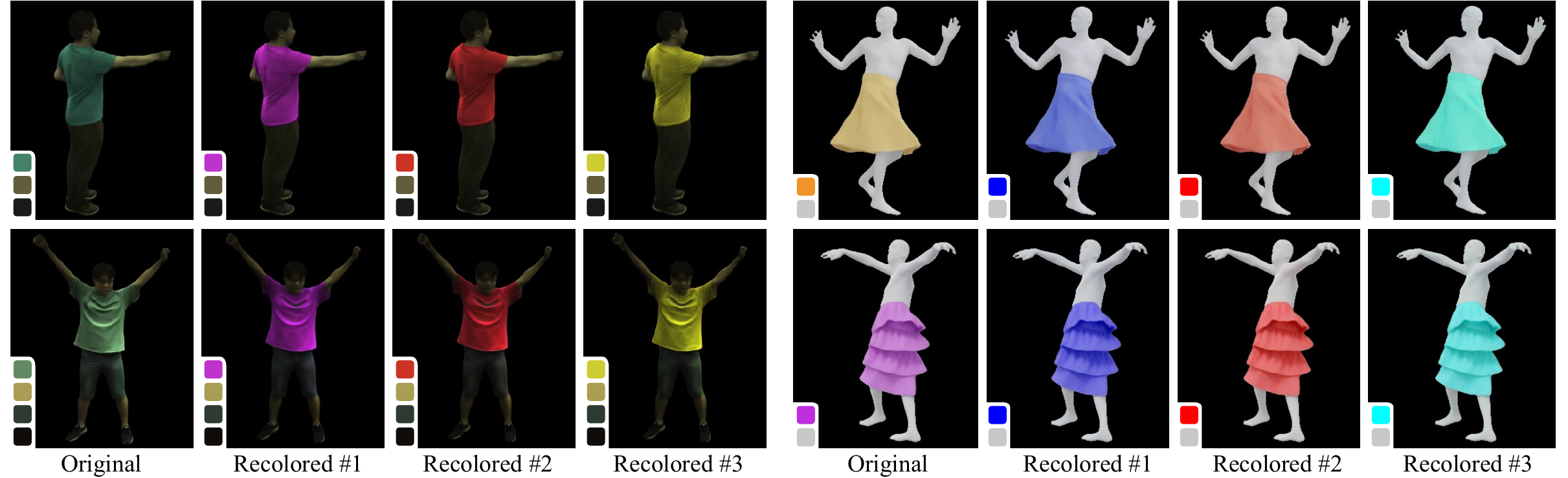}
\vskip -0.1in
\caption{\textbf{Color editing.} 
By adjusting the palette-based vectors, we can modify the color of the target garment appearance, while maintaining the fabric's fold details. }\label{fig:recolor}
\vskip -0.2in
\end{figure*}
%-------------------------------------------------

\textbf{Real-captured data.}
We demonstrate that our method is applicable to real-captured garment data. We utilized the multi-view rgb sequence provided by the ZJUMoCap database \cite{neuralbody}, along with the fitted SMPL body mesh for each frame. We train the network using 16 viewpoints and 700 frames from the video sequences, while the remaining frames were used to evaluate the model's generalization capability to the unseen motion. As illustrated in Figure \ref{fig:real_example}, our method is capable of generating plausible garment details and wrinkle effects across the unseen body motion.

\textbf{Color editing.}
we demonstrate the capability of color editing of our method. As discussed in Section \ref{sec: Rendering Network}, the palette-based decomposition of visual elements allows for easy manipulation of colors through adjusting the palette base color vector $p$. Figure \ref{fig:recolor} shows the recolored result on both synthetic and real capture data.
Compared to the rendering result in original color, the recolored results preserve the detailed folds and wrinkles without introducing artefacts.

% For each subject, the first column displays the original images, with the current palette base color $p$ displayed in the lower-left corner of each image. Our recolored results change the color of the garment without affecting the colors of other parts, while preserving the detailed folds of the fabric. 
% For real capture data (shown on the left side of Fig. \ref{fig:recolor}), we extract 3\~4 palette base colors to correspond with major elements such as tops, skin, pants, and hair. When the colors of pants and hair are identical, they are represented by a single palette base color.

\textbf{Computational performance.}
Once the training is complete, with our un-optimized python code, our network takes 174 ms per frame in total, to generate the animation of the character dressed in the target garment, including 130 ms to compute the information map $\{N_t, V_t\}$ of the moving body, 5 ms to generate reference images by running the image generative model $\mathcal{G}_{img}$, 39 ms to run the other modules of network. We conduct our experiments on a PC equipped with an AMD 5950X CPU, 64GB of memory, and an NVIDIA GeForce RTX 3090 graphics card.
%-------------------------------------------------
\begin{figure*}[t]
\centering
\includegraphics[width=.9\linewidth]{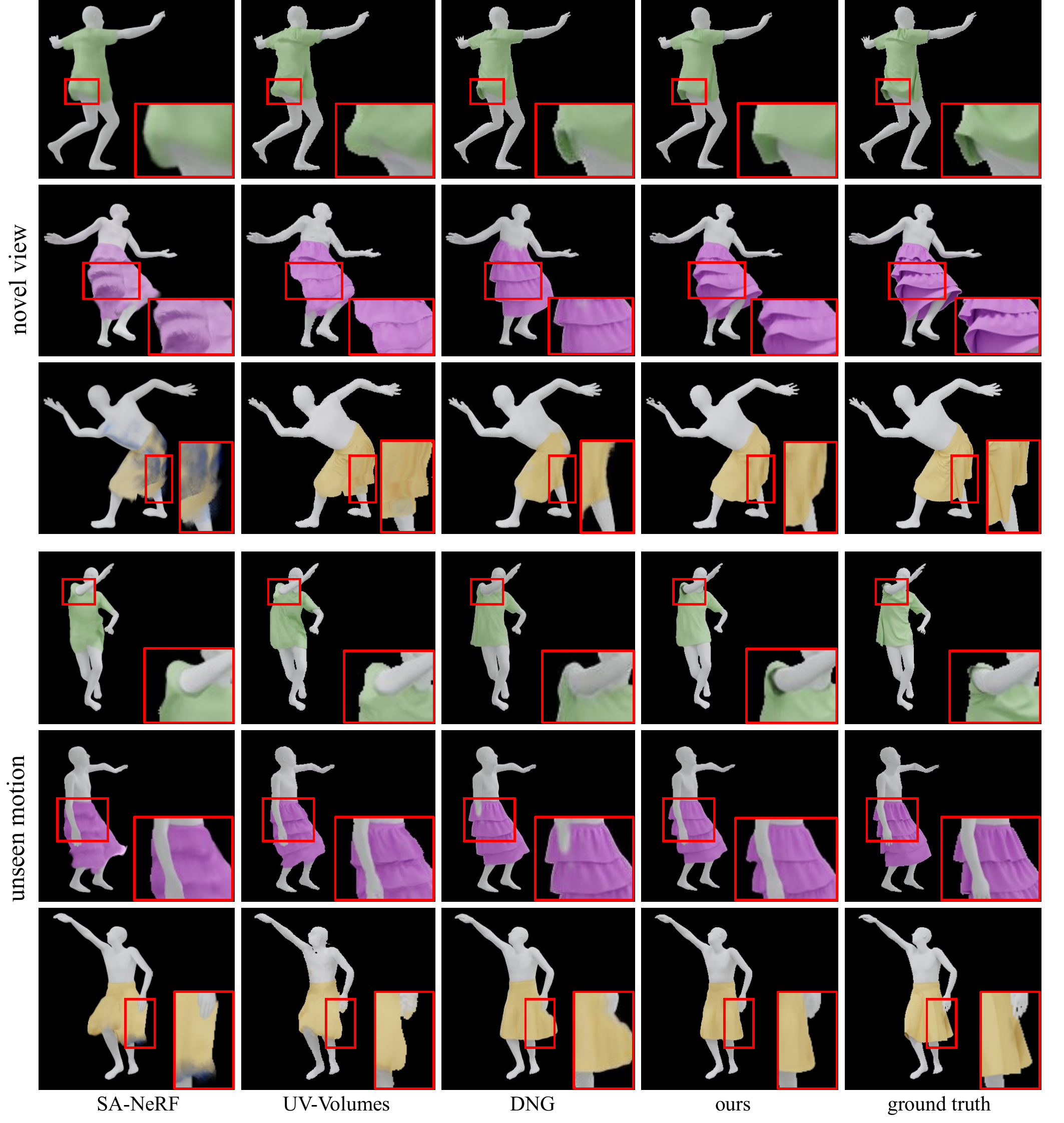}
\vskip -0.1in
\caption{\textbf{Comparisons.} We compare our method to SA-NeRF \cite{sanerf}, UV-Volumes \cite{chen2023uv}, as well as DNG \cite{dynamicneuralgarment},  on cases of both unseen camera views and body motions. Our approach outperforms these baseline methods.}\label{fig:compare}
\vskip -0.15in
\end{figure*}
%-------------------------------------------------

%-------------------------------------------------
\begin{figure}[t]
\centering
\includegraphics[width=1.\linewidth]{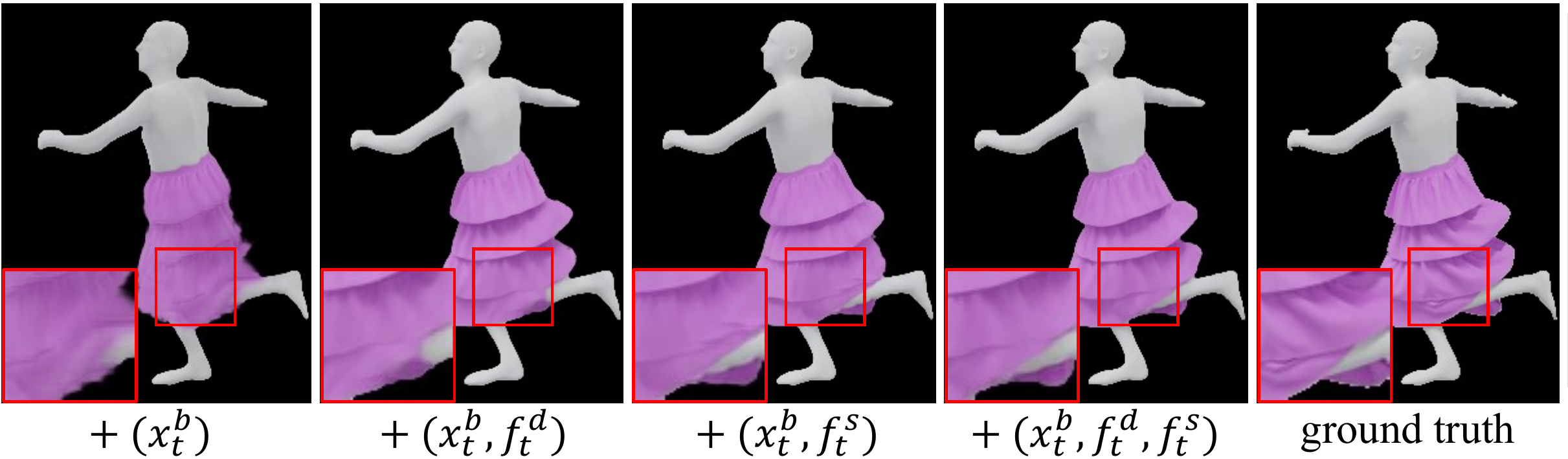}
\vskip -0.05in
\caption{The effect of different input features.}\label{fig:ablation_feat}
\vskip -0.15in
\end{figure}
%-------------------------------------------------
%-------------------------------------------------
\begin{figure}[t]
\centering
\includegraphics[width=1.\linewidth]{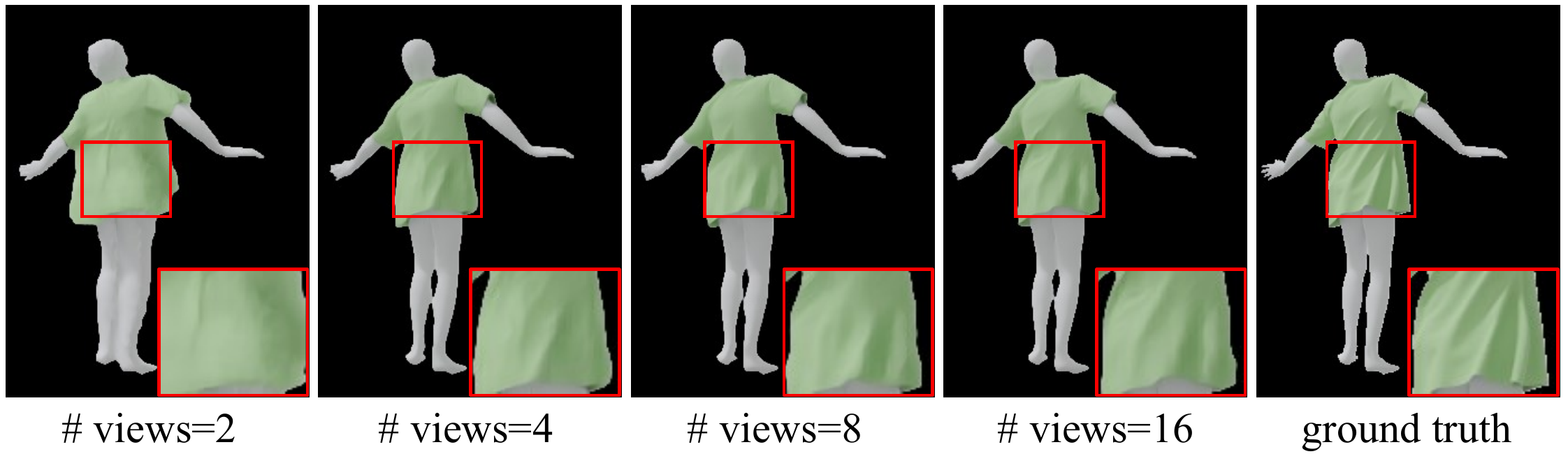}
\vskip -0.05in
\caption{The effect of number of sampled views during training.}\label{fig:ablation_views}
\vskip -0.15in
\end{figure}
%-------------------------------------------------

% Once the training is complete, our method requires 130 ms per frame for UV projection and an additional 4 ms for encoding. The image generator takes 5 ms to generate reference images and 30 ms for feature lookup, followed by 3 ms to encode the features. Finally, our rendering network takes 2 ms to render the final image. We conduct our experiments on a PC equipped with an AMD 5950X CPU, 64GB of memory, and an NVIDIA GeForce RTX 3090 graphics card.

% Table generated by Excel2LaTeX from sheet 'compare-v2 (2)'
% Table generated by Excel2LaTeX from sheet 'compare-v2 (2)'
% Table generated by Excel2LaTeX from sheet 'compare-v2 (2)'
\begin{table*}[t]\small
  \centering
    \begin{tabular}{|c|c|c|c|c|c|c|c|c|c|c|c|c|}
    \hline
    \multicolumn{13}{|c|}{seen motion} \\
    \hline
          & \multicolumn{4}{c|}{t-shirt}  & \multicolumn{4}{c|}{skirt}    & \multicolumn{4}{c|}{multilayer} \\
    \hline
    method & PSNR↑ & SSIM↑ & LPIPS↓ & tOF↓  & PSNR↑ & SSIM↑ & LPIPS↓ & tOF↓  & PSNR↑ & SSIM↑ & LPIPS↓ & tOF↓ \\
    \hline
    SA-NeRF & 23.121 & 0.933 & 0.136 & 0.716 & 21.510 & 0.915 & 0.166 & 0.877 & 22.085 & 0.912 & 0.147 & 0.766 \\
    UV-Volumes & 21.520 & 0.923 & 0.079 & 0.835 & 19.972 & 0.909 & 0.113 & 1.068 & 20.111 & 0.896 & 0.103 & 0.989 \\
    DNG   & 23.312 & 0.929 & 0.091 & 0.678 & 23.206 & 0.927 & 0.096 & 0.795 & 23.150 & 0.916 & 0.081 & 0.703 \\
    nearest view & 14.976 & 0.816 & 0.198 & 1.305 & 15.028 & 0.815 & 0.194 & 1.438 & 15.132 & 0.806 & 0.181 & 1.238 \\
    ours  & \textbf{27.737} & \textbf{0.974} & \textbf{0.044} & \textbf{0.520} & \textbf{26.446} & \textbf{0.966} & \textbf{0.055} & \textbf{0.681} & \textbf{26.677} & \textbf{0.960} & \textbf{0.050} & \textbf{0.578} \\
    \hline
    \multicolumn{13}{|c|}{unseen motion} \\
    \hline
          & \multicolumn{4}{c|}{t-shirt}  & \multicolumn{4}{c|}{skirt}    & \multicolumn{4}{c|}{multilayer} \\
    \hline
    method & PSNR↑ & SSIM↑ & LPIPS↓ & tOF↓  & PSNR↑ & SSIM↑ & LPIPS↓ & tOF↓  & PSNR↑ & SSIM↑ & LPIPS↓ & tOF↓ \\
    \hline
    SA-NeRF & 19.485 & 0.883 & 0.197 & 0.839 & 17.750 & 0.862 & 0.235 & 1.106 & 17.967 & 0.842 & 0.217 & 1.022 \\
    UV-Volumes & 18.822 & 0.877 & 0.127 & 1.073 & 17.568 & 0.862 & 0.158 & 1.341 & 17.401 & 0.837 & 0.151 & 1.323 \\
    DNG   & 20.405 & 0.893 & 0.140 & 0.771 & 19.500 & 0.883 & 0.152 & 0.967 & 19.562 & 0.860 & 0.133 & 0.893 \\
    ours  & \textbf{22.594} & \textbf{0.939} & \textbf{0.086} & \textbf{0.656} & \textbf{20.994} & \textbf{0.927} & \textbf{0.101} & \textbf{0.875} & \textbf{21.175} & \textbf{0.909} & \textbf{0.093} & \textbf{0.799} \\
    \hline
    \end{tabular}%
    \caption{To evaluate the generalization ability of our method and to compare with baseline methods, we show peak signal-to-noise ratio (PSNR) to qualify the image reconstruction quality of the synthesized images; structural similarity (SSIM) and perceptual similarity (LPIPS) between different methods and the ground truth image; and tOF \cite{chu2020learning} to measure the temporal consistency over time.
    % For quantitative comparisons, we present the PSNR, SSIM, LPIPS and tOF \cite{chu2020learning} scores between different methods and the ground truth image. The results indicate that our outcomes surpass those of the comparison methods across all metrics.
    }
  \label{tab:compare}%
  \vskip -0.15in
\end{table*}%

\subsection{Baseline Comparisons}
We compare our method against various baselines, including: (a) Surface Aligned Neural Radiance Fields (SA-NeRF) \cite{sanerf}, which constructs neural radiance fields based on relative positions to the human body surface; (b) the method of UV-Volumes \cite{chen2023uv}, which maps synthesized 2D textures to voxel latent spaces through predicted dense surface correspondences; and (c) Dynamic Neural Garment (DNG) \cite{dynamicneuralgarment}, which to synthesize the dynamic garment appearance based on learnable neural texture of a garment proxy.

We train the models of the baseline methods with the same training dataset as ours. 
% and were evaluated using the same movements from different viewpoints, as well as unseen motions, as depicted in Figure \ref{fig:compare}. 
The results in Figure \ref{fig:compare} demonstrate that although SA-NeRF and UV-Volumes have good performance in handling with the body-and-garment occlusions, they result in garment animation rendering with blurred wrinkle details and incorrect garment shape. Meanwhile, DNG yields visually appealing result but often produces incorrect occlusion relationships in areas like the hands and shoulders, and it can not ensure consistency of the detail structure in dense view rendering that we show the comparison in the supplemental video.
In comparison to those baseline methods, our method have better performance in terms of detail richness, contour accuracy, structural consistency and occlusion correctness. In Table \ref{tab:compare}, we conduct a quantitative comparison demonstrating that our method surpasses the comparative methods in both unseen camera viewpoints and body motions, and achieves higher quality of rendering results with better temporal consistency and more similarity to the ground truth animation.

%-------------------------------------------------
% Table generated by Excel2LaTeX from sheet 'compare-v2'
\begin{table*}[tbp]\small
  \centering
    \begin{tabular}{|c|c|c|c|c|c|c|c|c|c|}
    \hline
    \multicolumn{2}{|c|}{used features} & \multicolumn{4}{c|}{seen motion} & \multicolumn{4}{c|}{unseen motion} \\
    \hline
    detail feature & dynamic feature & PSNR↑ & SSIM↑ & LPIPS↓ & tOF↓  & PSNR↑ & SSIM↑ & LPIPS↓ & tOF↓ \\
    \hline
          &       & 22.160 & 0.925 & 0.112 & 0.765 & 20.590 & 0.901 & 0.130 & 0.844 \\
\hline 
\checkmark &       & 26.159 & 0.956 & 0.056 & 0.621 & 20.689 & 0.904 & 0.104 & 0.878 \\
\hline 
& \checkmark & 26.238 & 0.957 & 0.056 & \textbf{0.574} & 21.093 & 0.908 & 0.097 & \textbf{0.764} \\
\hline 
\checkmark & \checkmark & \textbf{26.677} & \textbf{0.960} & \textbf{0.050} & 0.578 & \textbf{21.175} & \textbf{0.909} & \textbf{0.093} & 0.799 \\
    \hline
    \end{tabular}%
    \caption{We present the quantitative evaluation of an ablation study for different input features. The results demonstrate that both detail feature and dynamic structural feature contribute to the improvement of our model's accuracy.}
  \label{tab:ablation_feat}%
  \vskip -0.1in
\end{table*}%
%-------------------------------------------------
%-------------------------------------------------
% Table generated by Excel2LaTeX from sheet 'compare-v2'
\begin{table*}[!tbp]\small
  \centering
    \begin{tabular}{|c|c|c|c|c|c|c|c|c|}
    \hline
          & \multicolumn{4}{c|}{seen motion} & \multicolumn{4}{c|}{unseen motion} \\
    \hline
    \# views & PSNR↑ & SSIM↑ & LPIPS↓ & tOF↓  & PSNR↑ & SSIM↑ & LPIPS↓ & tOF↓ \\
    \hline
    2     & 22.031 & 0.924 & 0.097 & 0.781 & 19.031 & 0.882 & 0.143 & 0.928 \\
\hline    4     & 26.717 & 0.969 & 0.049 & 0.561 & 22.399 & 0.938 & 0.088 & 0.689 \\
\hline    8     & 27.340 & 0.972 & 0.046 & 0.545 & 22.456 & 0.938 & 0.088 & 0.672 \\
\hline    16    & \textbf{27.737} & \textbf{0.974} & \textbf{0.044} & \textbf{0.520} & \textbf{22.594} & \textbf{0.939} & \textbf{0.086} & \textbf{0.656} \\
    \hline
    \end{tabular}%
    \caption{ We report PSNR, SSIM, LPIPS to study the effect of number of camera views during training on the quality of the synthesized image results. We show that training with a greater number of camera views enhances the generalization of our method across unseen camera views and body motions.
    % We present the quantitative evaluation results of the ablation study for different numbers of viewpoints. The results show that our model achieves satisfactory result with just four views and, as the number of views increases, its performance on unseen motion improves.
    }
  \label{tab:ablation_view}%
  \vskip -0.1in
\end{table*}%
%-------------------------------------------------
%-------------------------------------------------

\begin{table*}[t]\small
  \centering

    \begin{tabular}{|c|c|c|c|c|c|c|c|c|c|c|c|c|}
    \hline
    \multicolumn{13}{|c|}{seen motion} \\
    \hline
          & \multicolumn{4}{c|}{t-shirt}  & \multicolumn{4}{c|}{skirt}    & \multicolumn{4}{c|}{multilayer} \\
    \hline
    $k$     & PSNR↑ & SSIM↑ & LPIPS↓ & tOF↓  & PSNR↑ & SSIM↑ & LPIPS↓ & tOF↓  & PSNR↑ & SSIM↑ & LPIPS↓ & tOF↓ \\
    \hline
    0     & 27.451 & 0.973 & 0.046 & 0.530 & 26.458 & 0.966 & 0.058 & \textbf{0.680} & 26.380 & 0.959 & 0.051 & 0.584 \\
    1     & 27.522 & 0.973 & 0.045 & 0.541 & 26.555 & 0.966 & 0.058 & 0.691 & 26.493 & 0.959 & 0.053 & 0.588 \\
    2     & \textbf{27.737} & \textbf{0.974} & \textbf{0.044} & \textbf{0.520} & 26.446 & 0.966 & \textbf{0.055} & 0.681 & 26.677 & 0.960 & 0.050 & 0.578 \\
    5     & 27.725 & \textbf{0.974} & \textbf{0.044} & 0.525 & 26.587 & 0.966 & 0.057 & 0.688 & 26.686 & 0.960 & \textbf{0.049} & 0.583 \\
    10    & 27.592 & \textbf{0.974} & 0.045 & 0.537 & 26.533 & 0.966 & 0.057 & \textbf{0.680} & \textbf{26.704} & \textbf{0.961} & \textbf{0.049} & 0.577 \\
    20    & 26.636 & 0.967 & 0.055 & 0.683 & \textbf{26.636} & \textbf{0.967} & \textbf{0.055} & 0.683 & 26.593 & 0.960 & \textbf{0.049} & \textbf{0.570} \\
    \hline
    \multicolumn{13}{|c|}{unseen motion} \\
    \hline
          & \multicolumn{4}{c|}{t-shirt}  & \multicolumn{4}{c|}{skirt}    & \multicolumn{4}{c|}{multilayer} \\
    \hline
    $k$     & PSNR↑ & SSIM↑ & LPIPS↓ & tOF↓  & PSNR↑ & SSIM↑ & LPIPS↓ & tOF↓  & PSNR↑ & SSIM↑ & LPIPS↓ & tOF↓ \\
    \hline
    0     & 22.423 & \textbf{0.939} & 0.087 & 0.661 & 20.990 & 0.927 & 0.104 & 0.877 & 21.053 & 0.908 & 0.095 & 0.797 \\
    1     & 22.518 & \textbf{0.939} & 0.087 & 0.664 & \textbf{21.192} & \textbf{0.928} & \textbf{0.101} & 0.876 & 21.135 & \textbf{0.909} & 0.095 & 0.799 \\
    2     & \textbf{22.594} & \textbf{0.939} & \textbf{0.086} & 0.656 & 20.994 & 0.927 & \textbf{0.101} & 0.875 & \textbf{21.175} & \textbf{0.909} & \textbf{0.093} & 0.799 \\
    5     & 22.499 & \textbf{0.939} & \textbf{0.086} & 0.663 & 21.081 & \textbf{0.928} & \textbf{0.101} & 0.877 & 21.049 & 0.908 & 0.095 & 0.813 \\
    10    & 22.474 & 0.938 & 0.088 & 0.656 & 21.059 & 0.927 & 0.102 & \textbf{0.864} & 21.168 & \textbf{0.909} & \textbf{0.093} & 0.799 \\
    20    & 22.528 & \textbf{0.939} & \textbf{0.086} & \textbf{0.653} & 21.054 & 0.927 & 0.104 & 0.865 & 21.104 & 0.908 & 0.094 & \textbf{0.796} \\
    \hline
    \end{tabular}%
    \caption{    
    We quantitatively evaluate the impact of different historical frame numbers $k$. When $k=2$, our method demonstrates a significant generalization ability to unseen body motions, applicable to garment types of t-shirt, skirt and multi-layered dress.
    %The results show that when $k=2$, the model exhibits better generalization ability.
    }
     
  \label{tab:ablation_k}%
  \vskip -0.1in
\end{table*}%

%-------------------------------------------------
\subsection{Ablation Study}
\textbf{Different training features.}
We evaluate the effects of detail feature $f_t^d$ and dynamic structural feature $f_t^s$ on our model and present the results in Figure \ref{fig:ablation_feat} and Table \ref{tab:ablation_feat}. 
Our results show that models incorporating both detail and dynamic structural features outperforms those that include only the geometry feature ($x^b_t$), or the geometry feature combined with just one of the detail feature ($f_t^d$) and dynamic structure feature ($f_t^s$). We observe that integrating either the detail or dynamic feature alone can enhance the rendering quality, improving garment details or contour, respectively.

%As can be seen from Fig. \ref{fig:ablation_feat}, using either detail feature or dynamic structural feature alone can significantly improve the rendering quality， compared with the results only taking $x_t^b$ as the rendering network input. 

% We evaluate the effects of detail feature $f_t^d$ and dynamic structural feature $f_t^s$ on our model separately and presented the results in Fig. \ref{fig:ablation_feat} and Table \ref{tab:ablation_feat}. As can be seen from Fig. \ref{fig:ablation_feat}, using either detail feature or dynamic structural feature alone can significantly improve the rendering quality. However, a closer look at the red-boxed area in Fig. \ref{fig:ablation_feat} reveals two key observations: 
% First, using only the detail feature increases the visibility of wrinkles in the skirt, but because the detail feature itself is not structural consistent, it produces incorrect details at the intersection of two reference views $c^{fr}$ and $c^{ba}$, especially at the edges.
% Secondly, relying solely on the dynamic structural feature tends to produce a coherent geometry but lacks wrinkle details compared to the detail feature.
% By combining these two features, our final result not only obtained richer details, but also ensured geometric accuracy.

\textbf{Different number of training views.}
When training our Garment Animation NeRF, we set 16 camera poses to generate video of training data. In Figure \ref{fig:ablation_views} and Table \ref{tab:ablation_view}, we demonstrate the results of our model trained with varying numbers of views. Our results indicate that training with a greater number of camera views enhances the generalization of our method across unseen views and unseen body motion.  
Remarkably, our method can produce plausible rendering results when trained with animation data from only 4 camera views. However, because the primary focus of this paper is not on minimizing the number of views, we chose to train with 16 views in the remainder of our experiments to ensure the highest visual quality possible.

% Additionally, when the number of training views goes beyond 8, our method can result in comparable quality of garment animation rendering. 

% Our results show that our model is capable of rendering relatively stable and realistic results with just 4 views, which proves the strong convergence ability of our model. However, since the focus of this paper is not on reducing the number of views, we still opted for training with 16 views in the remainder of our experiments to ensure the best visual effects.

\textbf{Different number of historical frames.}
We quantitatively evaluate the impact of using different numbers of historical frames $k$ in the velocity map $V_t$ in Table \ref{tab:ablation_k}. 
When $k=0$ and only normal $N_t$ contributes to generate the dynamic feature $F_t^s$ with $V_t = 0$, our method produces plausible rendering results (with SSIM values above 0.9). 
The quality of rendering animation for seen motion sequences generally improves as more dynamic information introduced with increasing $k$.
However, for unseen motion sequences, the performance decreases when $k>2$. 
We speculate that while historical dynamic information aids in synthesizing garment animation, excessive historical information leads to over-fitting to the training motion sequences, weakening generalization to unseen body motions.
When $k=2$, our method demonstrates a significant generalization to unseen body motions, applicable to all garment types of t-shirt, skirt and multi-layered dress.
%Comprehensively considering the quantitative impact of different numbers of historical frames on t-shirt, skirt, and multi-layered dress in Table \ref{tab:ablation_k}, we set $k=2$ for other experiments to validate our method.   
%Our results show that when $k=2$, the model exhibits better generalization ability and achieves better results under unseen motion sequences.

\section{Limitations and future work}

\begin{figure}[!t]
\centering
\includegraphics[width=0.6\linewidth]{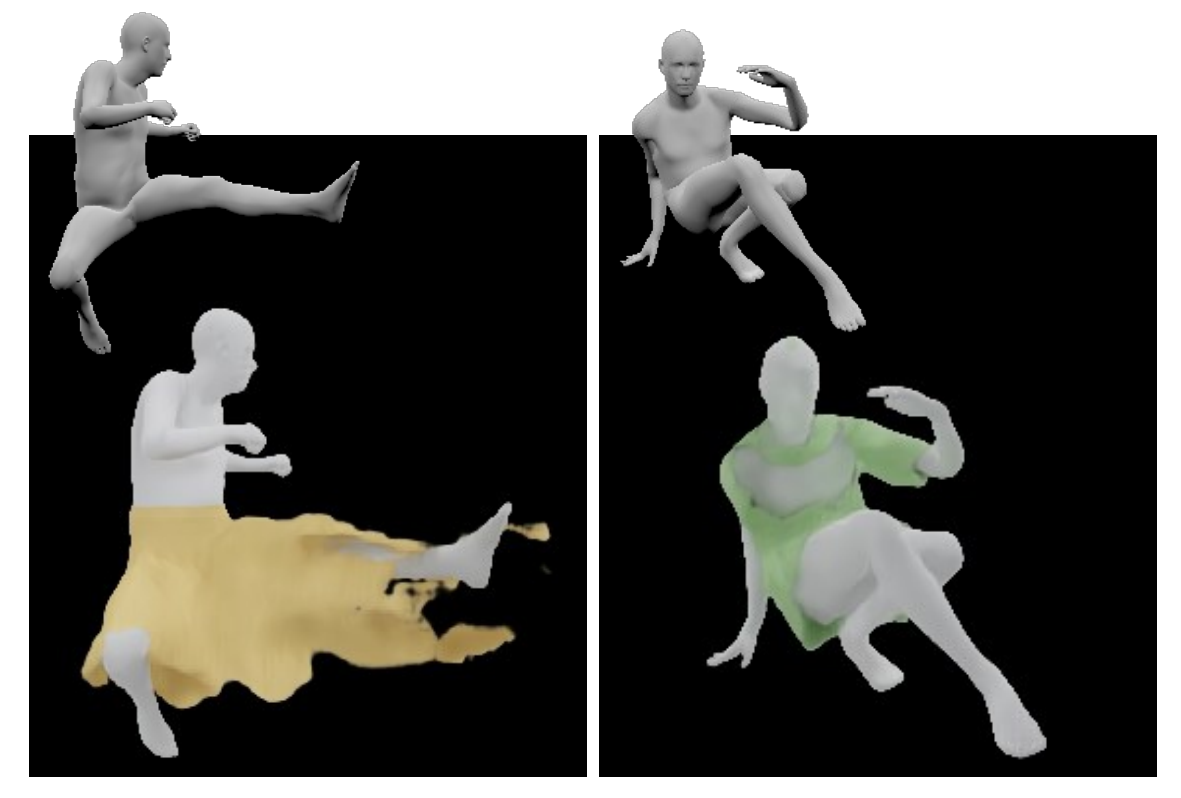}
\vskip -0.1in
\caption{
\textbf{Failure cases.} Our work falls short when synthesizing garment animation for unseen body motions distributed far away from the training data.
% Failure cases. When the input actions exceed the distribution of actions in our training set, unsatisfactory results can occur. Furthermore, as the gray-scale images $\hat{I}_t(c^{fr})$ and $\hat{I}_t(c^{ba})$ cannot ensure structural consistency, the information it provides may mislead our model. This can result in unstable geometry in the areas where projections overlap.
}\label{fig:bad_case}
\vskip -0.3in
\end{figure}

This work presents a novel framework for synthesizing garment animations, especially for complex and loose garments. 
We directly infer garment dynamics from the movement information of the body, eliminating the need for an explicit garment proxy. However, our work encounters limitations when synthesizing garment dynamics for unseen body motions distributed far away from those in the training set, as shown in Figure \ref{fig:bad_case}. To address this issue, expanding training dataset is a viable option. Additionally, a more intriguing direction might be to implicitly learn physically-realizable features integrated within the NeRF architecture.

%-------------------------------------------------
%\subsection{Limitation and future work.}

Regarding the capture of wrinkle details, we employ a generative image model to create reference appearance features. Given the rapid advancement in generative AI \cite{chen2020generative, martin2021nerf, poole2022dreamfusion} for producing realistic images and videos, we plan to further explore text-driven garment appearance editing in the future work. This exploration will involve generating detailed features tailored with pre-trained text-driven models, such as DALL-E \cite{ramesh2021zero, ramesh2022hierarchical} and Imagen \cite{saharia2022photorealistic}.  

 Currently, our work supports only color editing of the target garment by decomposing the visual elements of appearance features. Moving forward, we aim to focus more on disentangling the latent features of appearance, differentiating between garment style, texture patterns, materials. This will allow for more flexible and adaptive editing of garment appearances. 
 Moreover, accurately locating a desired 3D position within the implicit field of the dynamic garment constructed by NeRF is challenging. An promising direction for future work is to enable partial color editing of garment animation by transforming the implicit field into an explicitly accessible structural representation \cite{yang2022neumesh}.
 %\review{Moreover, guided by the spatial-and-temporal aware dynamic implicit field constructed by NeRF, an interesting direction for future work is to extend our work to enable partial color editing of garment animation.} 

%-------------------------------------------------
%-------------------------------------------------

 Furthermore, we believe the computational time efficiency of our method can be significantly improved in the future work. The most time-consuming steps are the information map recording (130ms) and feature lookup during the NeRF rendering process (30ms). We will try to use graph convolution network \cite{pfaff2020learning} to directly encode features based on the mesh topology of the body template, instead of running CNN on the 2D feature map. And meanwhile, we intend to incorporate data structures like octrees and hashtables \cite{garbin2021fastnerf, yu2021plenoctrees} into our garment animation NeRF, to more quickly locate relevant features during the rendering inference. It will be promising to enable real-time rendering of garment animation.

\section{Acknowledgments}
The authors would like to thank the reviewers for their constructive comments; Kun Wang and Xinxin Zou for their valuable suggestions. This work was partially supported by the National Science Fund of China, Grant Nos. 62072242 and 62361166670.

%-------------------------------------------------------------------------
% bibtex
\bibliographystyle{eg-alpha-doi} 
\bibliography{egbib}   

% biblatex with biber
% \printbibliography                

\end{document}

% uncomment for using teaser
% \teaser{
%  \includegraphics[width=0.9\linewidth]{eg_new}
%  \centering
%   \caption{New EG Logo}
% \label{fig:teaser}
%}

\maketitle
%-------------------------------------------------------------------------

\section{Symbol Table}

\begin{table*}[htbp]
  \centering
  \caption{Add caption}
    \begin{tabular}{|c|c|c|c|}
    \hline
    \multicolumn{1}{|c|}{Source} & \multicolumn{1}{c|}{Symbol} & Shape & describe \\
    \hline
    \multicolumn{1}{|c|}{\multirow{8}{*}{2D feature extractor}} &   $\mathcal{E}_{2D}$    & -     & 2D feature encoder \\
\cline{2-4}          &   $\mathcal{G}_{2D}$    & -     & 2D generator \\
\cline{2-4}          &   $\mathbf{c}_f$    & -     & Reference view (front) \\
\cline{2-4}          &   $\mathbf{c}_b$    & -     & Reference view (back) \\
\cline{2-4}          &   $\mathbf{I}_f$    & 512*512*3 & Reference image (front) \\
\cline{2-4}          &   $\mathbf{I}_b$    & 512*512*3 & Reference image (back) \\
\cline{2-4}          &   $\mathbf{F}_f$    & 512*512*128 & 2D feature map (front) \\
\cline{2-4}          &   $\mathbf{F}_b$    & 512*512*128 & 2D feature map (back) \\
    \hline
    \multicolumn{1}{|c|}{\multirow{5}{*}{3D feature extractor}} &  $\mathcal{E}_{3D}$     & -     & 3D feature encoder \\
\cline{2-4}          & t & 1 & Number of history frames \\
\cline{2-4}          &   $\mathbf{N}$    & 128*128*3 & Normal uv map \\
\cline{2-4}          &   $\mathbf{V}$    & 128*128*3t & Velocity uv map \\
\cline{2-4}          &   $\mathbf{F}_{3D}$    & 512*512*128 & 3D feature map \\
    \hline
    \multicolumn{1}{|c|}{\multirow{20}{*}{Renderer}} &   $\mathcal{M}$    & -     & MLP to predict $\mathbf{f}$ and $\sigma$ \\
\cline{2-4}          &   $\mathcal{D}$    & -     & 2D Neural renderer \\
\cline{2-4}          &   $\mathbf{x}$    & 3 & Spatial sample point \\
\cline{2-4}          &   $\mathbf{x}_p$    & 3 & $\mathbf{x}$’s nearest projection on mesh \\
\cline{2-4}          &   $\mathbf{x}_c$    & 3 & $\mathbf{x}_p$’s corresponding point on T-pose mesh \\
\cline{2-4}          &   $\mathbf{h}$    & 1 & L1 distance between $\mathbf{x}$ and $\mathbf{x}_p$ \\
\cline{2-4}          &  $\mathbf{f}_{2D}$     & 128 & 2D feature \\
\cline{2-4}          &  $\mathbf{f}_{3D}$     & 128 & 3D feature \\
\cline{2-4}          &  $\mathbf{f}$     & 128 & $\mathcal{M}$’s latent feature \\
\cline{2-4}          &  $\sigma$     & 1 & Density of radiance field \\
\cline{2-4}          &  $\mathbf{F}$ & 128*128*128 & Latent feature map after volume rendering \\
\cline{2-4}          &  $\mathbf{W}$     & 512*512*2*1 & Blend weight of palette colors \\
\cline{2-4}          &  $\mathbf{p}$     & 2*3   & palette colors \\
\cline{2-4}          &  $\mathbf{O}$     & 512*512*2*3 & Color offset \\
\cline{2-4}          &  $n_p$     & 1 & Number of palette colors \\
\cline{2-4}          &  $\pi_{vol}$     & -     & Volume rendering \\
\cline{2-4}          &  $\mathbf{M}_{p}$     & 512*512*1 & Predicted front mask \\
\cline{2-4}          &  $\mathbf{M}_{gt}$     & 512*512*1 & Ground truth front mask \\
\cline{2-4}          &  $\mathbf{I}_{p}$     & 512*512*3 & Predict image \\
\cline{2-4}          &  $\mathbf{I}_{gt}$     & 512*512*3 & Ground truth image \\
    \hline
    \end{tabular}%
  \label{tab:addlabel}%
\end{table*}%

%-------------------------------------------------------------------------
% bibtex
\bibliographystyle{eg-alpha-doi} 
\bibliography{egbib}   

% biblatex with biber
% \printbibliography                